\documentclass[lettersize,journal]{IEEEtran}
\usepackage{amsmath}
\usepackage{amsfonts}
\usepackage{algorithmic}
\usepackage{algorithm}
\usepackage{array}
\usepackage[caption=false,font=normalsize,labelfont=sf,textfont=sf]{subfig}
\usepackage{textcomp}
\usepackage{stfloats}
\usepackage{url}
\usepackage{verbatim}
\usepackage{graphicx}
\usepackage{cite}
\usepackage{amssymb}

\usepackage{color}
\begin{document}

\title{Riemannian Complex Matrix Convolution Network for PolSAR Image Classification}

\author{Junfei Shi,~\IEEEmembership{IEEE member}, Wei Wang, Haiyan Jin, Mengmeng Nie,Shanshan Ji \vspace{-2em}

\thanks{Junfei Shi was with  the Department
of Computer Science and Technology, Shaanxi Key Laboratory for Network Computing and Security Technology, Xi'an University of Technology, Xi'an, China. Corresponding author: shijunfei@xaut.edu.cn.}
\thanks{Wei Wang, Haiyan Jin, Mengmeng Nie and Shanshan Ji were with the Department
of Computer Science and Technology, Shaanxi Key Laboratory for Network Computing and Security Technology, Xi'an University of Technology, Xi'an, China.}}


\markboth{Journal of \LaTeX\ Class Files,~Vol.~14, No.~8, July~2023}%
{Shell \MakeLowercase{\textit{et al.}}: A Sample Article Using IEEEtran.cls for IEEE Journals}


\maketitle

\begin{abstract}
Recently, deep learning methods have achieved superior performance for Polarimetric Synthetic Aperture Radar(PolSAR) image classification. Existing deep learning methods learn PolSAR data by converting the covariance matrix into a feature vector or complex-valued vector as the input. However, all these methods cannot learn the structure of complex matrix directly and destroy the channel correlation. To learn geometric structure of complex matrix, we propose a Riemannian complex matrix convolution network for PolSAR image classification in Riemannian space for the first time, which directly utilizes the complex matrix as the network input and defines the Riemannian operations to learn complex matrix's features. The proposed Riemannian complex matrix convolution network considers PolSAR complex matrix endowed in Riemannian manifold, and defines a series of new Riemannian convolution, ReLu  and LogEig operations in Riemannian space, which breaks through the Euclidean constraint of conventional networks. Then, a CNN module is appended to enhance contextual Riemannian features. Besides, a fast kernel learning method is developed for the proposed method to learn class-specific features and reduce the computation time effectively. Experiments are conducted on three sets of real PolSAR data with different bands and sensors. Experiments results demonstrates the proposed method can obtain superior performance than the state-of-the-art methods.
\end{abstract}


\section{Introduction}

\IEEEPARstart{P}{olarimetric} synthetic aperture radar (PolSAR) system\cite{9906133,9143461} can acquire scattering echoes by emitting and receiving electromagnetic waves with two sets of antennas based on horizontal and vertical polarimetric bases. Thus, for each resolution unit, a scattering matrix can be acquired instead of a back-scattering coefficient, which can provide more scattering mechanism and polarimetic information for ground objects. PolSAR system can obtain rich scattering information in day-night and all-weather, which has been widely used in object detection\cite{9885221}, agriculture monitoring\cite{li2023unsupervised}, disaster prediction\cite{8888558} and land cover classification\cite{10092884} etc. Among them, PolSAR image classification\cite{8994163} is the indispensable preliminary for PolSAR image interpretation and understanding, which can assign each pixel to a certain class, such as urban area, farmland, water, mountain etc.

  In decade years, many conventional PolSAR image classification methods\cite{9269434,673687,8899856} have been proposed to make full use of polarimetric information of PolSAR data, which can be summarized as three categories. The first one is target decomposition-based methods, including H/$\alpha$/A decomposition\cite{9906133}, Freeman and Durden decomposition\cite{673687}, four decomposition\cite{8899856}, polarimetric coefficient extraction\cite{8946891} etc. These methods attempt to extract different scattering features to distinguish various ground objects. However, these pixel-wise methods are easily affected by speckle noises. The second one is the statistical distribution-based methods, in which the PolSAR covariance matrix is modeled as Wishart\cite{9106405}, mixed Wishart\cite{8438543}, K\cite{7730236}, G0\cite{6228506} and Kummer U\cite{6891291} distributions respectively. Different distributions can model various types of land covers, while no one can describe all kinds of terrain objects. The third one is computer vision-based methods, in which both scattering and image features are extracted, followed by feature learning and classifier design. These methods can take advantages of image features, while they only extract low-level features without considering high-level semantic.

   Recently, deep learning-based methods\cite{shi2017,8771134,8518517} have attracted widespread attention and make remarkable achievement in various application fields\cite{xie2020polsar}, such as image classification, segmentation, object detection and recognition etc. Deep learning methods can learn high-level features automatically and improve classification performance for PolSAR images, in which the CNN model\cite{8771134,he2016deep} is the mainstream. However, for each resolution unit, PolSAR data is a complex covariance matrix instead a vector or a scalar value. The complex-valued matrix cannot be directly fed into a conventional deep network as the input. To learn deep polarimetric information, the common processing approach is to convert the $3\times3$ complex matrix into a 9-dimension vector\cite{8558689} as the input of the network, or extracting a set of scattering features\cite{9172110} as the input vector. Then, deep features are learned for classification. For instance, Wang et. al\cite{8518517} proposes a multi-layer convolution LSTM model to learn the polariemtric features well. Bi et al.\cite{8486693} proposes a graph-based semisupervised deep model to learn both the high-level features and boundaries. Liu et al.\cite{8558689} gives the polarimetric convolutional neural network, which attempts to learn polarimetric scattering information by coding the real and imaginary parts of complex matrix as a new vector format. Hua et al.\cite{9172110} proposes a three-channel CNN, which can learn different scattering features respectively. Shang et al\cite{shang2020dense} gives the dense connection and depthwise separable network, which can independently extract features over each channel in PolSAR images to improve the classification accuracy. Zeng et al\cite{9624979} proposes a double-encoder network, which joint encodes the polariemtric and phase features to segment PolSAR images adaptively. All these methods utilize polarimetric feature vector as the input, and try to learn deep features by designing various networks. However, for SAR imaging system, both the amplitude and phase are the important scattering coefficients, which have not been fully explored by using the real-valued vector, since the imagery part of the complex backscattering coefficient stands for the phase information.

   To fully explore the amplitude and phase information, the complex-valued CNN network\cite{zhang2017complex} is proposed, in which the computing unit is the complex-valued data instead of the scalar value. This model is a complex-valued end-to-end network, in which the amplitude and phase can be learned simultaneously. It is the pioneering work for PolSAR network from real to complex data domain. Based on the complex-valued CNN network, multiple of complex-valued deep models have been proposed for PolSAR image classification. For instance, Tan et al\cite{8864110} proposes the complex-valued 3-D CNN network, which can learn 3-D image block in complex domain. Li et al\cite{li2020complex} gives the complex contourlet-CNN network, which can learn both spatial and structural information in complex frequency domain. In addition, Zhang et al\cite{9319226} combines the complex-valued 3D-CNN and hybrid condition random field (CRF) to learn the contextual information. Tan et al\cite{9345443} gives the deep triple complex-valued network, which uses three branches to learn complex data from different channels respectively. Later, Tan et al\cite{9771283} proposes the complex-valued variational inference network, which adds the variational inference into the complex-valued network to improve the classification accuracy. Furthermore, a complex-valued LSTM network\cite{9694655} is proposed, which can learn complementary information from multi-dimension features by LSTM network, as well as adding the phase information to improve classification performance. Jiang et al\cite{jiang2022unsupervised} proposes the complex-valued sparse feature learning method to perform the feature learning with the small size of samples. In a word, these complex-valued networks covert the complex covariance matrix into complex-valued vector to learn high-level features, which can learn phase information effectively. However, complex-valued network destroys the matrix structure and cannot learn the channel correlation for PolSAR data. So, it is deeply thought whether we can learn the complex matrix directly and construct a complex matrix network to learn the matrix structure and channel correlation for PolSAR data.

   Based on the thought above, it is repeatedly considered that how to develop a complex matrix network to learn polarimetric covariance matrix. It is found that the covariance matrix is a Hermitian positive definite (HPD) matrix for each resolution unit in the PolSAR image, which endows in a HPD manifold in Riemannian space instead of Euclidean space\cite{7947120,9963700,2009Polarimetric}. However, conventional deep learning methods include the convolution, ReLu and pooling operations, which are designed in Euclidean space. All these operations are not suitable for PolSAR complex matrix network anymore. New Riemannian convolution, ReLu and pooling operations in Riemannian space should be developed for PolSAR complex matrices. So, a new Riemannian complex matrix network is necessary to be established.

   To learn the geometric structure of complex matrices, some manifold learning methods\cite{chen2020covariance,8545822} in Riemannian space should be helpful for constructing the Riemannian complex matrix network. To be specific, the Riemannian metric of two HPD matrix should be the geometric distance in manifold space. The common used Riemannian metrics are the affine invariant Riemannian metric (AIRM), Stein, Jeffrey and log Euclidian distances \cite{chen2020covariance,8545822} etc. Based on these metrics, some classification methods have been proposed for PolSAR image classification. For instance, Zhong et al\cite{7947120} proposes Riemannian sparse coding for PolSAR image classification. Shi et al\cite{9963700} proposes Riemannian nearest regularization subspace method, which utilizes the Riemannian distance and infers a new optimization method. Later, Shi et al\cite{9884297} develops the complex matrix and multi-feature collaborative learning method by defining two kinds of dictionaries and designs different measures to improve classification performance.

   Recently, some deep learning methods in Riemannian space have been developed, and are initially applied to computer vision tasks. For example, Huang et al\cite{huang2017riemannian} proposes the SPDnet, which uses the Symmetric Positive Definite(SPD) matrix as the computation unit in the deep network in manifold space. Wang et al\cite{9390301} gives the lightweight network to reduce computation cost in SPDnet. Zhang et al\cite{8960323} proposes a transfer network from manifold to manifold to improve the ability of action recognition. Wang et al\cite{9043722} gives a multi-kernel metric learning method to measure the similarity of multiple Riemannian manifold data. These methods give the preliminary study of manifold network on Riemannian space. However, these methods are based on real-valued matrices and only tested on some simple visual tasks, which have not considered the complex-valued matrix information and scattering characteristics of PolSAR images. At present, there is still no relative work on the complex matrix network for PolSAR data in Riemannian space.

  To learn polarimetric complex matrix directly in deep network, we propose the Riemannian complex matrix convolution network (RCM\_CNN) for PolSAR image classification for the first time. The main contributions can be summarized as three aspects.
\begin{enumerate}
\item[1)]A Riemannian complex matrix network backbone (RCMnet) is proposed, which is the end-to-end complex matrix network in Riemannian space. It is the first work to give a manifold to manifold complex matrix network for PolSAR image classification, which breaks through the constraint of deep model in Euclidian space. The proposed RCMnet backbone redefines the Riemannian convolution, ReLu and LogEig operations with manifold metrics in Riemannian space, in which the complex matrix is the computation unit instead of a vector.

\item[2)]The RCMnet can learn the geometric structure of complex matrix well in Riemannian space. However, it is the pixel-wise feature learning method without considering contextual information. To learn contextual high-level features, a Riemannian complex matrix convolutional network (RCM\_CNN) is put forward by connecting a CNN network to the tail of the RCMnet, by which the features are converted from Riemannian to Euclidian space, and a softmax classifier is used for classification. Experimental results verify the effectiveness of the proposed RCM\_CNN method.

\item[3)]To reduce the computation cost, a fast polarimetric kernel learning method is proposed, in which the convolution kernels are calculated by the $(2D)^2PCA$ method, which is a fast interpretable network without back-propagation to project pixels within class together. Thus, the proposed RCMnet can learn the class-specific Riemannian features.
\end{enumerate}

 The remainder of this paper is organized as follows. Section II introduces the preliminary of PolSAR data and Riemannian manifold. Section III details the proposed RCM\_CNN network, including the Riemannian convolution, ReLu and pooling operations in the RCM backbone, and the convolution kernel learning method. In Section IV, the experimental results and analysis are illustrated to verify the effectiveness of the proposed method. Finally, the brief conclusion and further work are discussed in Section V.

\section{Preliminary}
In this section, the PolSAR data and Riemannian metrices are introduced in detail.
\subsection{PolSAR data}
Under the horizontal and vertical polarimetric base, the PolSAR data can be expressed as a scattering matrix S for each scattering unit, defined as:
\begin{equation}\label{ee1}
  S = \left[ {\begin{array}{*{20}{c}}
{{S_{hh}}}&{{S_{hv}}}\\
{{S_{vh}}}&{{S_{vv}}}
\end{array}} \right]
\end{equation}
where $S_{hh}$ is the scattering signal of horizontal emitting and vertical receiving polarization.  $S_{hv}$, $S_{vh}$ and $S_{vv}$ have the similar conception. In the monostatic radar case, $S_{hv}=S_{vh}$. So,  the scattering matrix can be vectored under the Pauli basis as follows.

\begin{equation}\label{ee2}
k = \frac{1}{{\sqrt 2 }}{\left[ {{S_{hh}} + {S_{vv}},{S_{hh}} - {S_{vv}},2{S_{hv}}} \right]^T}
\end{equation}

In general, multi-look processing is necessary for despeckling of PolSAR data. After processing, the coherency matrix is acquired, which is the most common used PolSAR data representation defined as:

\begin{equation}\label{ee3}
T = \frac{1}{N}\sum\limits_{i = 1}^N {{k_i}k_i^T}  = \left[ {\begin{array}{*{20}{c}}
{{T_{11}}}&{{T_{12}}}&{{T_{13}}}\\
{{T_{21}}}&{{T_{22}}}&{{T_{23}}}\\
{{T_{31}}}&{{T_{32}}}&{{T_{33}}}
\end{array}} \right]
\end{equation}

\subsection{Riemannian metric}

It is widely known that the polarimetric coherency matrices are Hermitian  positive definite (HPD), which forms a Riemannian manifold instead of Euclidean space\cite{2009Polarimetric,7947120}. The similarity of two points in Riemannian manifolds are measured by the Riemannian metric, which can be extended to PolSAR data space. There are four well-known metric distances in Riemannian space. One is the affine invariant Riemannian metric (AIRM), and another is log-Euclidean Riemannian metric (LERM). The third one is Jensen\_Bregman Logdet divergence or Stein divergence, and the last one is Jeffrey divergence.

1)AIRM: For two points in PolSAR image, the geometric distance of two HPD matrices X and Y can be defined as:

\begin{equation}\label{T1}
{d_R}\left( {X,Y} \right) = {\left\| {\log \left( {{X^{ - \frac{1}{2}}}Y{X^{ - \frac{1}{2}}}} \right)} \right\|_F}
\end{equation}

2)the log-Euclidean Riemannian metric (LEM) is defined as:
\begin{equation}\label{t2}
{d_L}\left( {X,Y} \right) = {\left\| {\log \left( X \right) - \log \left( Y \right)} \right\|_F}
\end{equation}

3)Stein divergence is defined as:
\begin{equation}\label{t3}
d_S^2\left( {X,Y} \right) = \log \det \left( {\frac{{X + Y}}{2}} \right) - \frac{1}{2}\log \det \left( {XY} \right)
\end{equation}

4)Jeffrey divergence is defined as:
\begin{equation}\label{t4}
d_J^2\left( {X,Y} \right) = \frac{1}{2}Tr\left( {{X^{ - 1}}Y} \right) + \frac{1}{2}Tr\left( {{Y^{ - 1}}X} \right) - n
\end{equation}

All the four Riemannian metrics induce a Riamannian geometry. Among them, AIRM is affine invariant in a curved geometry space. However, it is computationally expensive. LERM flattens the manifold by mapping into the tangent space, which is rotation and scale invariant separately. Stein divergence, also known as Jensen\_Bregman Logdet divergence, is a statistical-based similarity measure based on Bregman divergences with less computational cost. Furthermore, it shares similar properties as AIRM, e.g., invariant to affine transformation. These distances are defined for SPD matrices, and their geometry naturally extends to the HPD case\cite{7565529}.

\section{Proposed Method}
This section gives the framework of the proposed Riemannian complex matrix convolution network (RCM\_CNN) for PolSAR images as shown in Fig.1.
The proposed RCM\_CNN network consists of two modules. One is the proposed Riemannian complex matrix network module (RCMnet), which can learn complex matrix features in Riemannian space. The other is the CNN network module for high-level feature learning. The backbone of RCM network includes the Riemannian convolution and nonlinear rectifying in each hidden layer, ending with a log-mapping and fully connection layers. It is an end-to-end network architecture in Riemannian space. This module explores a manifold complex-valued matrix network in Riemannian space to learn matrix channel correlation. Then, the Riemannian features are converted into Euclidean space by log-mapping operation. The proposed RCMnet is the pixel-wise feature learning in Riemannian space, which ignores the contextual information. In the second module, a CNN network is appended on the tail of the RCMnet to learn the contextual information of Riemannian features. According to the Riemannian features learned from the RCMnet module, the CNN network is added to further learn the high-level contextual features. The architecture of the Riemannian complex matrix convolution network (RCM\_CNN) is designed in Section III-A. Then, a polarimetric kernel learning method is proposed to learn the network weights without back-propagation, which is described in Section III-B.

\begin{figure*}
\centering
\setlength{\fboxrule}{0.2pt}
  \setlength{\fboxsep}{0.01mm}
\includegraphics[height=0.2\textheight]{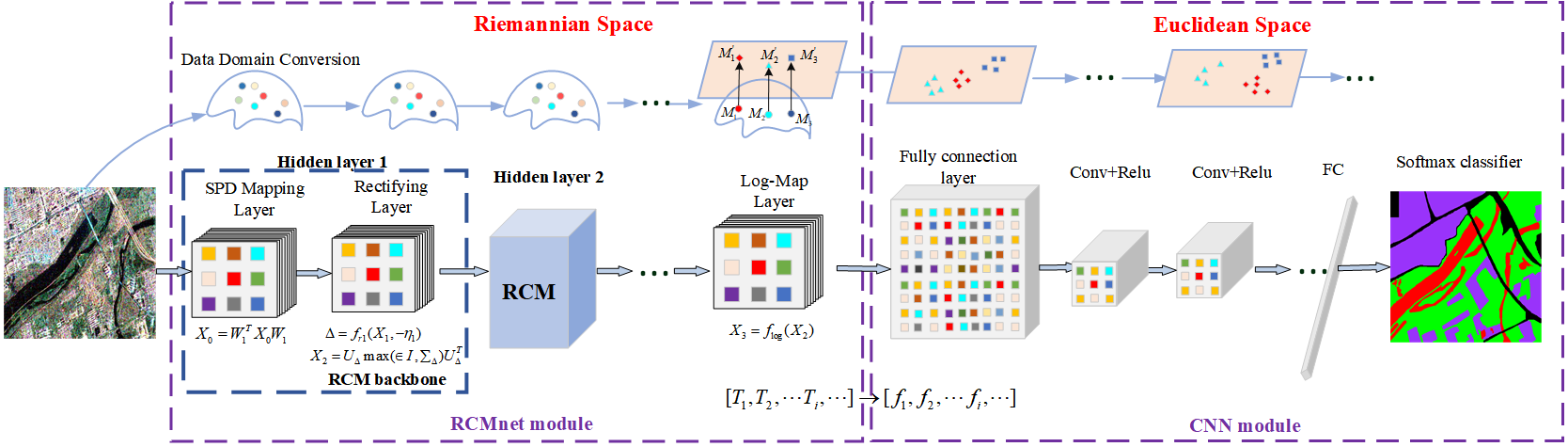}
\caption{Framework of the proposed Riemannian complex matrix convolution network.}
\label{fig1}
\end{figure*}

\subsection{Architecture of the Riemannian Complex Matrix Convolutional Network}

The architecture of the proposed RCM\_CNN network is given in Fig.1, which consists of the proposed RCMnet and CNN models. First of all, the RCMnet is a manifold to manifold network architecture in Riemannian space, including the Riemannian complex matrix mapping and nonlinear rectifying in each hidden layer, and ending with the log-mapping and fully connection layers. The input of the network is the original PolSAR coherency matrix. The PolSAR coherency matrix T is the complex-valued HPD matrix, which lies in Riemannian space. The traditional network cannot learn the structure of complex matrix with Euclidean measurement. So, a novel complex-valued matrix network architecture is necessary for PolSAR complex matrix learning endowed in Riemannian manifold.

The framework of the proposed RCMnet method is similar to the traditional ConvNet. The RCMnet module includes several hidden layers to learn the Remmainian features of complex matrices, which can preserve manifold structure well in Remmainian space. Then, the full connected layer is designed to flatten the Remainnian features into tangent space, in which Euclidean operations can be applied. Thus, features is transform from Riemannian to Euclidean space, so that the CNN enhanced layers can be utilized to learn high-level features. Finally, the softmax classification layer is added for classification. In each hidden layers, there are the convolution, nonlinear mapping (such as ReLu, sigmoid functions) and LogEig operations. However, in the CNN, these operations are calculated by Euclidean measurement for real-valued images. For PolSAR image, each pixel can be represented by a complex-valued T matrix, which is HPD matrix in Riemannian space. Therefore, the RCM network is proposed, in which each element should be fully complex-valued matrix information with Riemannian metric, including the filters, complex matrix (CM) mapping, nonlinear rectifying and LogEig layer operations.


The details of each element in RCM\_CNN network are presented as follows.

\emph{1)Riemannian complex matrix mapping layer}

Inspired by the effectiveness of ConvNet in learning the high-level features, we try to learn the discriminating features and class-specific information of PolSAR data. However, for PolSAR images, each pixel is presented as the coherency/covariance matrix, which is HPD matrix with Riemannian geometrical structure. So, traditional convolutional operation in Euclidean space is not suitable for PolSAR data at all. In the proposed RCM network, we give the Riemannian complex matrix mapping, which can learn the polarimetric complex matrix structure and geometric features of PolSAR data.

Our target is to learn the class-specific geometrical discriminating features from T matrix. To preserve matrix's geometrical structure in the new feature space, the resulting matrix should also be Hermiden positive definiteness. So, we design the Riemannian complex matrix mapping by using the bilinear mapping function $f_b$ as follows, which can transform the input HPD matrix to a new HPD matrix.

\begin{equation}\label{e1}
{X_k} = {f_b}\left( {{X_{k - 1}},{W_k}} \right) = W_k^T{X_{k - 1}}{W_k}
\end{equation}
where ${X_{k - 1}}$ is the output of the $k-1$th layer, and the initial input is the original T matrix when $k=1$. ${W_k}$ is the filter kernel, which is the connected weights in the network. ${X_k}$ is the resulting matrix after transformation. To ensure ${X_k}$ is a valid HPD matrix, ${W_k}$ is required as the row full-rank matrix\cite{9390301}. PolSAR matrix is rich in the polarimetric scattering information and multi-channel correlation. To learn the polarimetric class discriminating information in the new feature space, we select ${W_k}$ as the $3\times3$  class central complex matrix, which can be resolved by the proposed polarimetric kernel learning method in Section III-B. This method is a fast and effective kernel learning method without back-propagation. Thus, the CM mapping can convert the original data to a new feature space which enhances the class information and preserves the matrix manifold.

\emph{2)Riemannian Rectifying layer}

Rectified linear unit (ReLU) is the commonly used nonlinear activation function in the ConvNet, which can greatly improve the discriminating features by rectifying the results. Here, we want to extend the nonlinear rectifying layer to the complex matrix network in Riemannian space. Inspired by the ReLU function (including the max(0,x) non-linearity), we devise the nonlinear function $fr$ to rectify the result from the mapping layer. This function is calculated by eigenvalue decomposition. Then, the small eigenvalues are rectified by a threshold. So, this rectifying layer is named the ReEig rectifying layer. $fr$ is defined as:

\begin{equation}\label{e2}
{X_k} = f_r^{(k)}({X_{k-1}}) = {U_{k - 1}}\max \left( {\varepsilon I,{\Sigma _{k - 1}}} \right)U_{k - 1}^T
\end{equation}
where ${X_{k-1}} = {U_{k-1}}{\Sigma _{k-1}}U_{k-1}^T$, ${U_{k-1}}$ and ${\Sigma _{k - 1}}$ are the eigenvector and eigenvalue matrix by  singular value decomposition(SVD) of ${X_{k-1}}$. In matrix ${\Sigma _{k - 1}}$, the diagonal elements are eigenvalues, and others are zeros. $\max \left( {\varepsilon I,{\Sigma _{k - 1}}} \right)$ is the operation which replaces the small eigenvalues to $\varepsilon$ as follows:
\begin{equation}\label{e3}
\max \left( {\varepsilon I,{\Sigma _{k - 1}}} \right) = \left\{ {\begin{array}{*{20}{c}}
   {{\Sigma _{k - 1}}(i,i),{\rm{  }}{\Sigma _{k - 1}}(i,i) > \varepsilon ,}  \\
   {\varepsilon ,{\rm{             }}{\Sigma _{k - 1}}(i,i) < \varepsilon .}  \\
\end{array}} \right.
\end{equation}

Here, we set $\varepsilon $ above the smallest eigenvalue. Thus, the ReEig layer will always rectify the mapping matrix with nonlinear mapping, even all the eigenvalues are much greater than zero. This operation ensures the resulting matrices are positive definite.

Based on the non-linear design above, the Riemannian complex matrix mapping layer and ReEig rectifying layers are considered as a hidden unit in the RCM network. This hidden unit can learn the manifold geometric structure of complex matrices, and discriminating features can be achieved after multi-level hidden units. Besides, the transformed matrices are still HPD, which makes sure the scalability and stability of the RCM network structure.

\emph{3)LogEig layer}

In the end of the ConvNet, the full connected layer is used to integrate all the features from the upper layer. This layer can provide effective features for the final classified layer. Nevertheless, in the proposed method, the learned features are complex matrices after multi-level hidden units in the Riemannian manifold. So, to flatten the HPD manifold space, the logarithm operation has been proved to convert the manifold data into the corresponding tangent space, such that the Euclidian measurement can be utilized. Instead, the exponential operation (exp(.)) can map the point in tangent space into corresponding manifold space. According to this theory, we use complex matrix logarithm operation (logm(.)) to covert a manifold complex matrix to the flat tangent space, named the LogEig layer. Then, the Euclidean computation can be applied to the tangent space\cite{2007Geometric}. Here, the LogEig function is defined as:

\begin{equation}\label{e4}
\begin{array}{l}
 {X_k} = {f_{\log }}\left( {{X_{k - 1}}} \right) = \log \left( {{X_{k - 1}}} \right) \\
 {\rm{     }} = {U_{k - 1}}diag\left( {\log \left( {{\Sigma _{k - 1}}} \right)} \right)U_{k - 1}^T \\
 \end{array}
\end{equation}
where ${X_{k - 1}} = {U_{k - 1}}{\Sigma _{k - 1}}U_{k - 1}^T$ is the eigenvalue decomposition. $diag\left( {\log \left( {{\Sigma _{k - 1}}} \right)} \right)$
is the diagonal matrix of eigenvalue logarithm operation.

\emph{4)Fully connection layer}

In the LogEig layer, we have converted the Remminan CM data into the flat space, such that Euclidean computations can be applied to the domain of HPD matrix logarithm. To facilitate the subsequent classification, the learned CM features in the LogEig layer should be vectored. Since each CM is always with the size of $3\times3$, we transform the $i$th matrix into a 9-dimension feature vector, and fully connect all the CM features into a long feature vector $f \in {R^{9{n_k} \times 1}}$, where ${n_k}$ is the number of hidden unit in the $k$th layer. Thus, each PolSAR CM ${T_i}$ can be converted into a full connected feature vector ${f_i}$ after the RCM network. That is
\begin{equation}\label{e5}
[{T_1},{T_2}, \cdots {T_i}, \cdots ] \to [{f_1},{f_2}, \cdots {f_i}, \cdots ]
\end{equation}

In a word, the HPD complex matrix in Remminan space can be fully learned by the proposed RCM network, and final mapped into the feature vector $\textbf{F}$ in Euclidean space.

\emph{5) CNN-based feature enhancement}

After RCM network, the \textbf{T} matrix is converted into the feature vector $\textbf{F}$. However, it is a pixel-wise feature learning procedure. To learn higher-level feature and obtain contextual information, a CNN enhanced module is designed based on the output of the RCM module. In this module, $n$-layer CNN is utilized to achieve contextual features. After CNN enhanced module, Remianian feature $\textbf{F}$ is converted into high-level features $\textbf{H}$ as follows.

\begin{equation}\label{e6}
[{f_1},{f_2}, \cdots {f_i}, \cdots ] \to [{h_1},{h_2}, \cdots {h_i}, \cdots ]
\end{equation}

\emph{6) Softmax classifier}

For the learned feature vector $h$, the softmax classifier is used for PolSAR image classification. The trained samples are randomly selected from each class, and the cross-entropy loss is used, defined as:

\begin{equation}\label{e7}
{L_i} =  - \sum\limits_{c = 1}^M {{y_c}\log ({p_c})}
\end{equation}
where ${y_c}$ is the label of sample $i$, ${L_i}$ is the loss function of sample $i$. ${p_c}$ is the probability of sample $i$ belong to class $c$, which is defined as:

\begin{equation}\label{e8}
{p_c} = P({y_i} = c|{h_i}) = \frac{{{e^{h_i^T{w_c}}}}}{{\sum\limits_{k = 1}^C {{e^{h_i^T{w_k}}}} }}
\end{equation}

\subsection{Fast Polarimetric Kernel Learning Method}

For RCMnet, the backpropagation is a procedure of resolving complex matrix operations iteratively. During the network training, the computation of convolution kernel is the most time-costed, since some matrix operations are needed such as SVD decomposition and matrix inverse operation\cite{huang2017riemannian}. To reduce the calculating complexity, we propose the polarimetric kernel learning method to learn the weights of filters. This method can calculate the filter weights directly without backpropagation by utilizing $(2D)^2PCA$ method\cite{2DPCA,9390301}. Besides, each filter is calculated from the class center of coherency matrix by the Frechet mean. The principal component feature vectors are extracted and formulated the kernel matrix for each class. By this way, we can find a better projection space for each class, in which pixels within the same class are projected together, and the class-specific features can be learned.

$(2D)^2PCA$ Algorithm: $(2D)^2PCA$ method can reduce a $m\times n$ dimension matrix by using the row and column projection matrices X and Y respectively. It can be expressed as:
\begin{equation}\label{e91}
\tilde T = {X^T}TY
\end{equation}
To learn polarimetric kernel filters by the $(2D)^2PCA$ algorithm, for all training samples in each class, the coherency  matrix T are considered as the basic samples. Then, the sample covariance matrix is calculated from the row direction is expressed as

\begin{equation}\label{e9}
C_c = \frac{1}{{N - 1}}\sum\limits_{i = 1}^N {{{\left( {{T^i} - \bar T} \right)}^T}\left( {{T^i} - \bar T} \right)}
\end{equation}
where $\bar T$ is the mean of all the basic samples for class $c$. The mean of coherency matrix can be calculated by the Frechet mean in Riemannian space. The Frechet mean\cite{9390301} is defined as:

\begin{equation}\label{e10}
{\bar T}^* = \mathop {\arg \min }\limits_{{\bar T} \in S_{ +  + }^{d*}} \sum\limits_{i = 1}^N {{D_{LEM}}\left( {T_i^k,{\bar T}} \right)}
\end{equation}
where $N$ is the pixel number in class $k$, and $\bar T$ is the Frechet mean to be learn, known as the Reminanian barycenter. ${C_i^k}$ is the $i$th covariance matrix in class $k$. According to equation (\ref{e10}), the Frechet mean can be achieved by:

\begin{equation}\label{e11}
{\bar T}^* = \exp \left[ {\frac{1}{N}\sum\limits_i^N {\log \left( {T_i^k} \right)} } \right]
\end{equation}

According to Frechet mean, the sample covariance matrix can be calculated by equation (\ref{e9}). Then, the minimization of the reconstruction error defined in the $(2D)^2PCA$ is utilized to learn the target transformation matrix as follows:

\begin{equation}\label{e12}
\mathop {\min }\limits_{X \in {R^{3*3}}} \sum\limits_{i = 1}^N {\left\| {{T_i} - X{X^T}{T_i}} \right\|} _F^2,{\rm{  s}}{\rm{.t}}{\rm{. }}X{X^T} = I
\end{equation}
where \textbf{I} is the identity matrix. It has been proved that the solution of equation (\ref{e12}) is composed of a family of eigenvectors corresponding to the P largest eigenvalues of sample covariance matrix $C_c$ \cite{2DPCA}. Since the convolutional kernel is the same size with $3\times3$. Therefore, the $c$th polariemtric convolutional kernel is obtained by the three eigenvectors corresponding to the 3 largest eigenvalues of $C_c$.

%
The proposed RCM\_CNN algorithm can learn the geometric structure of complex matrix in Riemannian space by RCMnet module, and after feature translation, it can further learn the high-level features by the CNN model in Euclidean space. The algorithm procedure of the proposed RCM\_CNN method is given in Table \ref{t1}.

\begin{table*}[ht]
\footnotesize
\begin{center}
\caption
{\label{t1}
Algorithm procedure of the proposed RCM\_CNN classification method}
\begin{tabular}{p{16cm}}
\hline
\textbf{Algorithm 1} Proposed RCM\_CNN Classification Method\\
\hline
\textbf{Input: }PolSAR complex matrix \textbf{T} and the corresponding label map \textbf{L}. The number of classes $C$.\\

\emph{Step 1:} Input the coherency matrix T into the RCMnet module, and computing the kernel filters by using equations (\ref{e9}) and (\ref{e12}).\\
\emph{Step 2:} obtain the Riemannian feature matrices $X$ from RCMnet module, and translate the Riemannian feature into Euclidean feature $F$ after full connection layer. \\
\emph{Step 3:} Conduct the CNN module to enhance the  feature  $F$ into the higher-level features $H$.\\
\emph{Step 4:} Conduct the softmax classification and obtain the class label estimation map \textbf{Y}.\\
\textbf{Output:} class label estimation map \textbf{Y}.\\
\hline
\end{tabular}
\end{center}
\end{table*}

\section{Experimental Results and Analysis}

\subsection{Experimental data and settings}

 In this paper, three real PolSAR datasets are used to verify the effectiveness of the proposed methods. They are commonly used PolSAR datasets with different bands and sensors, which are described as follows.

A)\emph{\textbf{Oberpfaffenhofen data set}}: The first data set is the L-band full polarimetric SAR image over Oberpfanffenhofen, Germany area, acquired by E-SAR sensor from the German Aerospace Center. This image consists of  $1300\times 1200$ pixels with the resolution of $3\times2.2$m. It mainly covers five types of terrain objects, including \emph{urban area}, \emph{forest}, \emph{road}, \emph{farmland} and \emph{open area}. The PauliRGB image and its ground truth map are given in Figs.\ref{fig2}(a)-(b) respectively.

B)\emph{\textbf{Xi'an data set}}: The second one is the Xi'an dataset covering the Wei River in Xi'an, Shaanxi, China. It is acquired by Rararset-2 sensor with the Fine Quad Polarization model in January 2010. A subimage is used with the size of $512\times512$ pixels in our experiment. There are mainly three classes, including \emph{buildings}, \emph{grass} and \emph{water}. The PauliRGB image and its ground truth map are given in Figs. \ref{fig4}(a)-(b) respectively. It is noted that the black region in the ground truth map is the void region without label.

C)\emph{\textbf{San Francisco data set}}: The third full polariemtric SAR dataset is the C-band four-look data set acquired by RADARSAT-2 sensor over San Francisco area. This PolSAR image is $1800\times1380$ pixels with the resolution of 5$m$. This image is widely used for PolSAR image classification since it covers both natural and man-made targets, including \emph{water}, \emph{vegetation}, \emph{high-density}, \emph{low-density} and \emph{developed} urban areas. The PauliRGB image and its ground truth map are presented in Figs.\ref{fig6}(a)-(b) respectively.


In addition, four existing methods are used for comparison to test the effectiveness of the proposed method. They are described as follows in detail.

1)Super\_RF\cite{9664519}:The first one is the superpixel and polarimetric features based classification method (shorted by "Super\_RF"), in which polarimetric feature-based random forest method is utilized to obtain pixel-wise classification result, and then superpixels are used to correct noisy classes. In this model, multiple of polarimetric features are extracted as the input of the random forest. However, the proposed method and other compared methods only use the original data as the input.

2)CVCNN\cite{zhang2017complex}: The second one is the complex-valued CNN method (noted by "CVCNN"), which defines a complex-valued CNN network to learn the phase information well. In the CVCNN method, the coherency matrix T is converted to a complex-valued vector as the input of the network. The network structure includes five convolution layers, a mean pooling layer and a full connection layer.

3)DFGCN\cite{liu2020deep}: The third method is the deep fuzzy GCN method (noted by "DFGCN"), which develops the hybrid metric and the fuzzy weighted graph convolution network to learn the contextual information. This method is graph-based GCN method, which is different from CNN-based methods. In this method, the wishart distance is calculated, and the coherency matrix T is utilized as the input. The weight decay coefficient is $5 \times {10^{{\rm{ - }}6}}$, and the drop rate is 0.5.

4)MPCNN\cite{cui2021polarimetric}: The last one is the recent proposed polarimetric multi-path CNN method (noted by "MPCNN") for PolSAR image classification, which includes multiple paths to learn polarimetric features. In this method, the input is the column vector which is also acquired from T matrix. In each path, the basis network is PolCNN, which consists of five convolution layers, two max-pooling layers and a full connection layer.

In addition, the 9D-CNN method is utilized as the ablation study to test the effectiveness of the RCMnet module. In this method, the T matrix is converted into a 9-dimension column vector as the input. The network structure contains five convolution layers, two max-pooling layers and one full connection layer. The convolution kernel sizes are $5\times5$, $3\times3$, $3\times3$, $3\times3$, $1\times1$ respectively. This network structure is the same as the proposed method. The unique difference is that we replace the first conventional convolution layer by the single layer RCM model in the proposed method, which can learn Riemannian features effectively. In this paper, the proposed method uses one layer of the RCM module, on the tail with the CNN model(RCM1\_CNN), since experimental results indicate the single layer of RCM module has similar performance with less computation cost below.
Besides, to quantitatively evaluate the proposed method, several evaluation indicators are calculated, including per-class accuracy, overall accuracy (OA), average accuracy(AA), Kappa coefficient and confusion matrix. It is noteworthy that pixels in the black region are not taken into account during calculating the accuracy since they are unlabeled void region.

In addition, the network parameters are given as follows. The learning rate is 0.005, and the iteration number is set as 400. The Adam optimizer is selected in the network. The sample proportion of training samples is 10\%. Training samples are randomly selected from the labeled data. The patch size is set as $13\times13$. All the experimental results are the average values of running 5 times. The experiments are conducted on Windows 10 system with Matlab 2016a, with a computer of Intel Core i7CPU and 64 RAM. Besides, the experiments are conducted with NVIDIA GeForce RTX 3060 GPU with 12G memory.

\subsection{Experimental results on Oberpfaffenhofen data set}

\begin{figure*}
\centering
\setlength{\fboxrule}{0.2pt}
  \setlength{\fboxsep}{0.01mm}
\subfloat[]{\includegraphics[height=0.14\textheight]{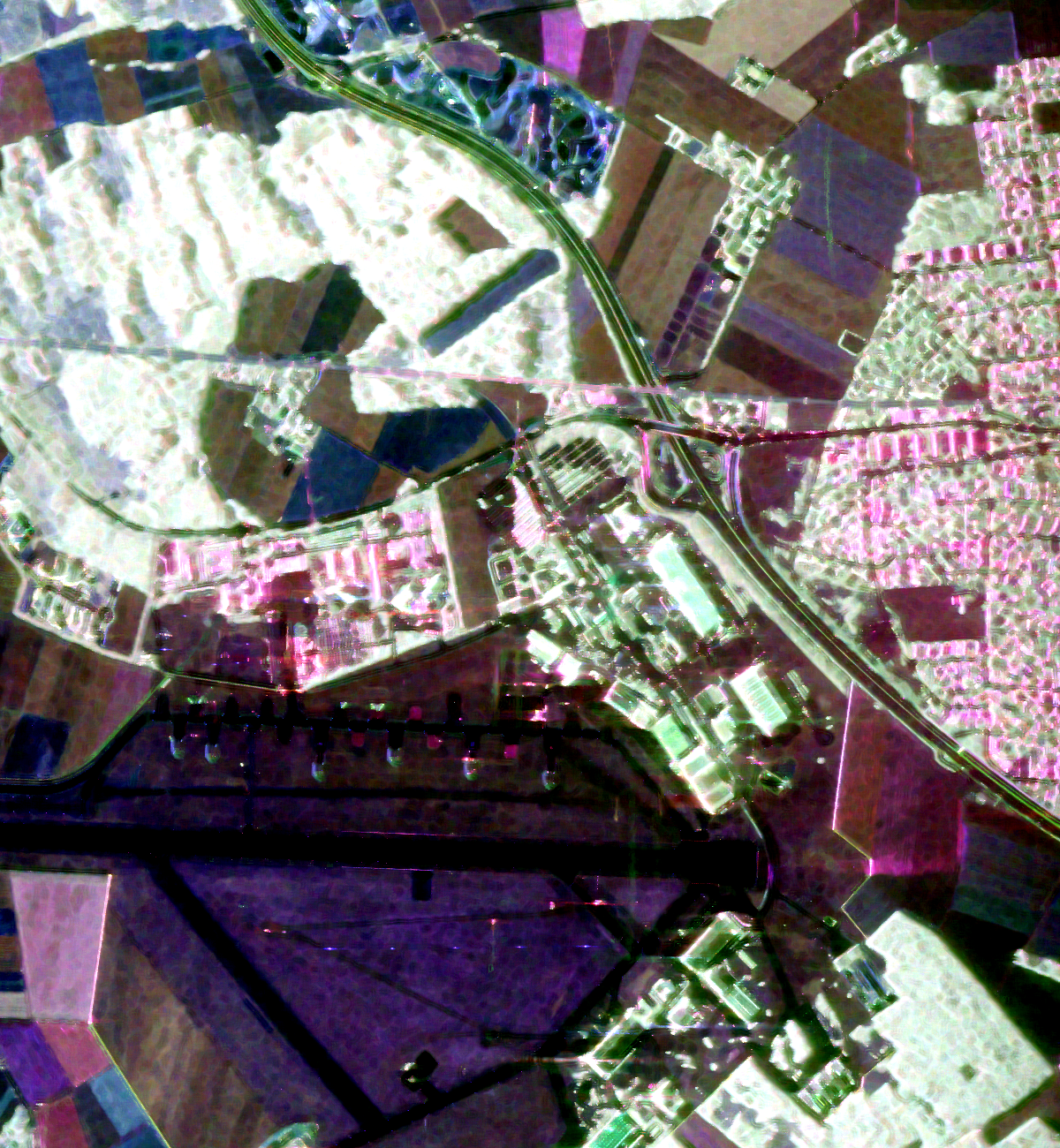}}
\subfloat[]{\fbox{\includegraphics[height=0.14\textheight]{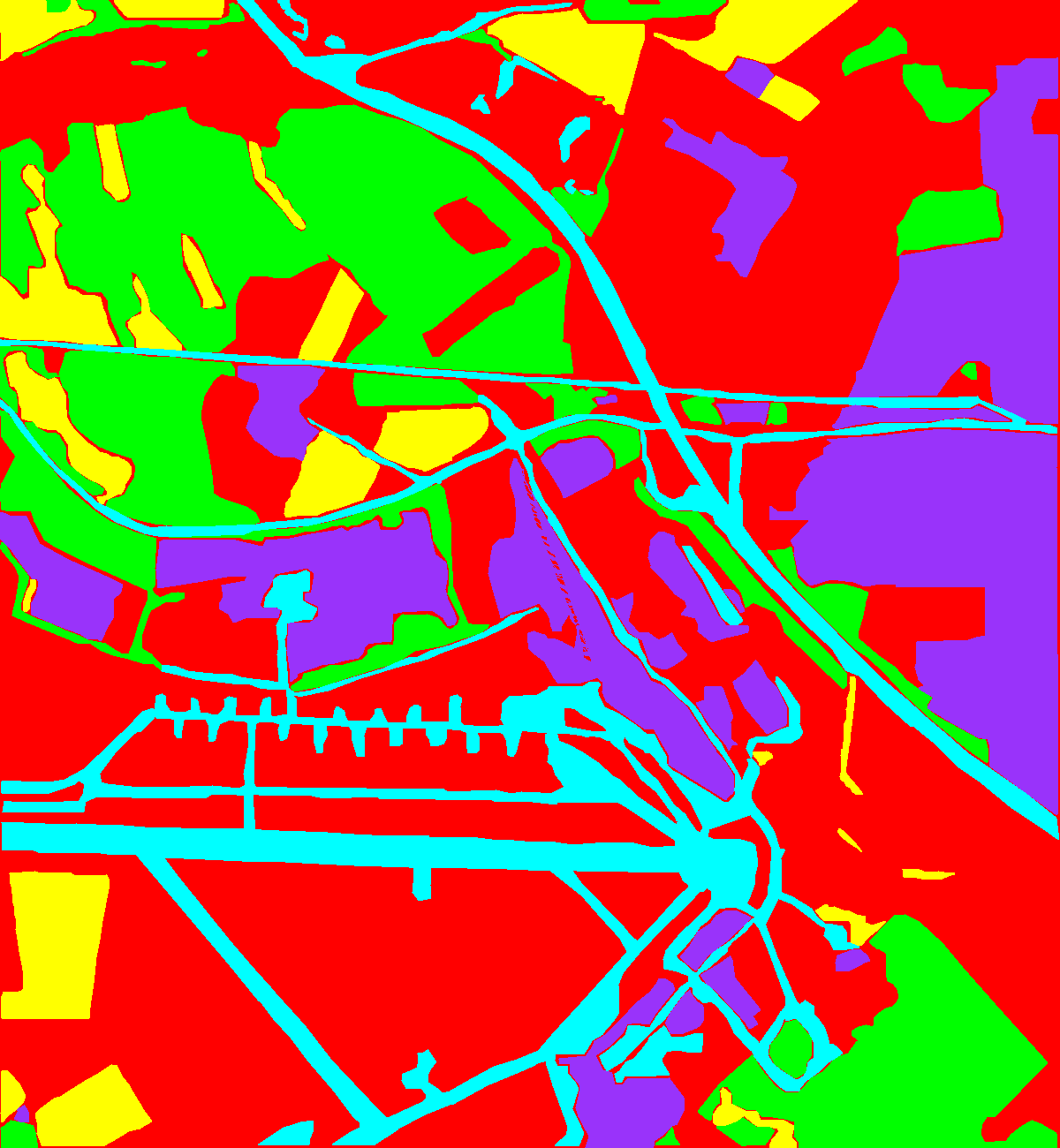}}}
\subfloat[]{\fbox{\includegraphics[height=0.14\textheight]{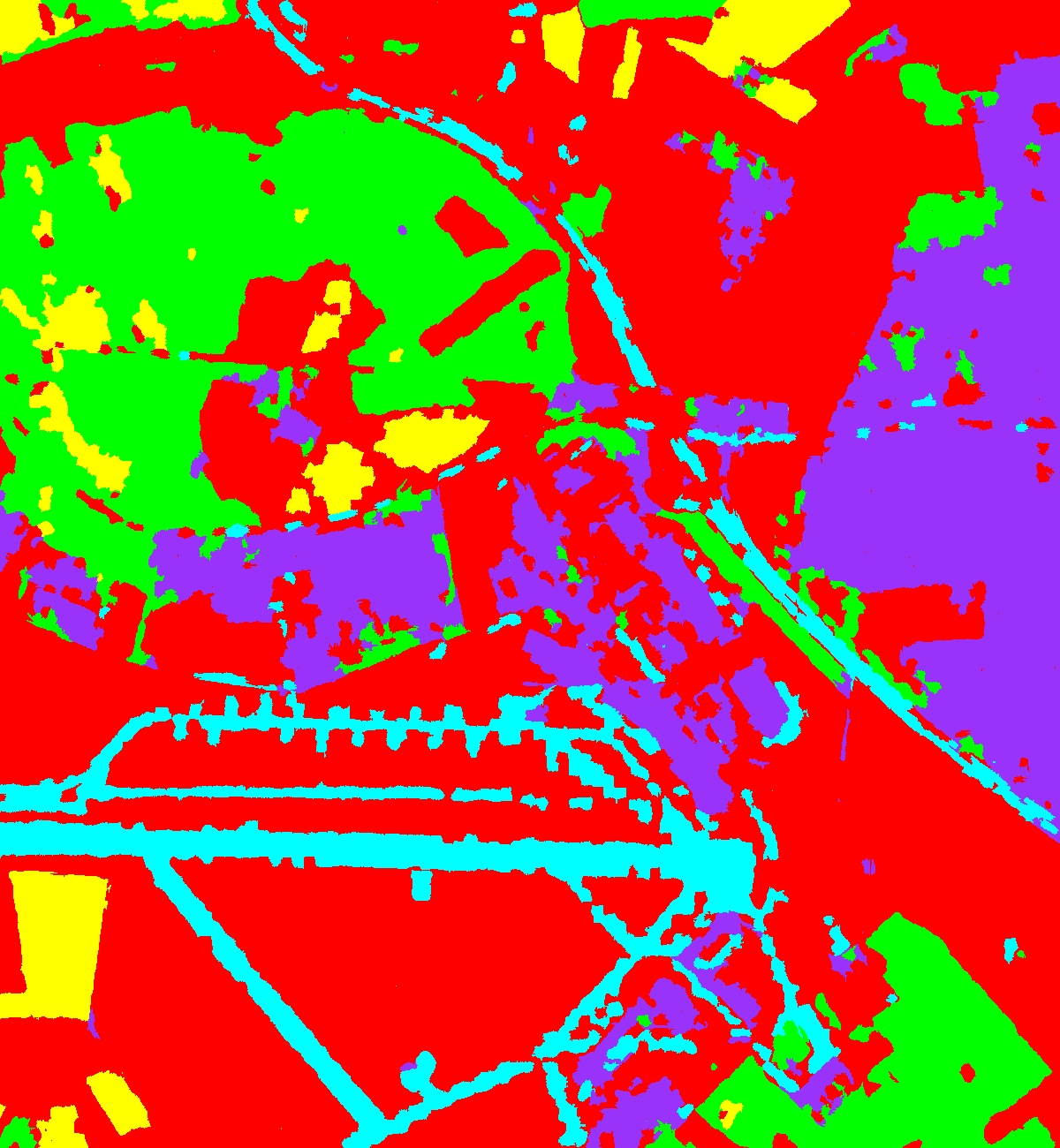}}}
\subfloat[]{\fbox{\includegraphics[height=0.14\textheight]{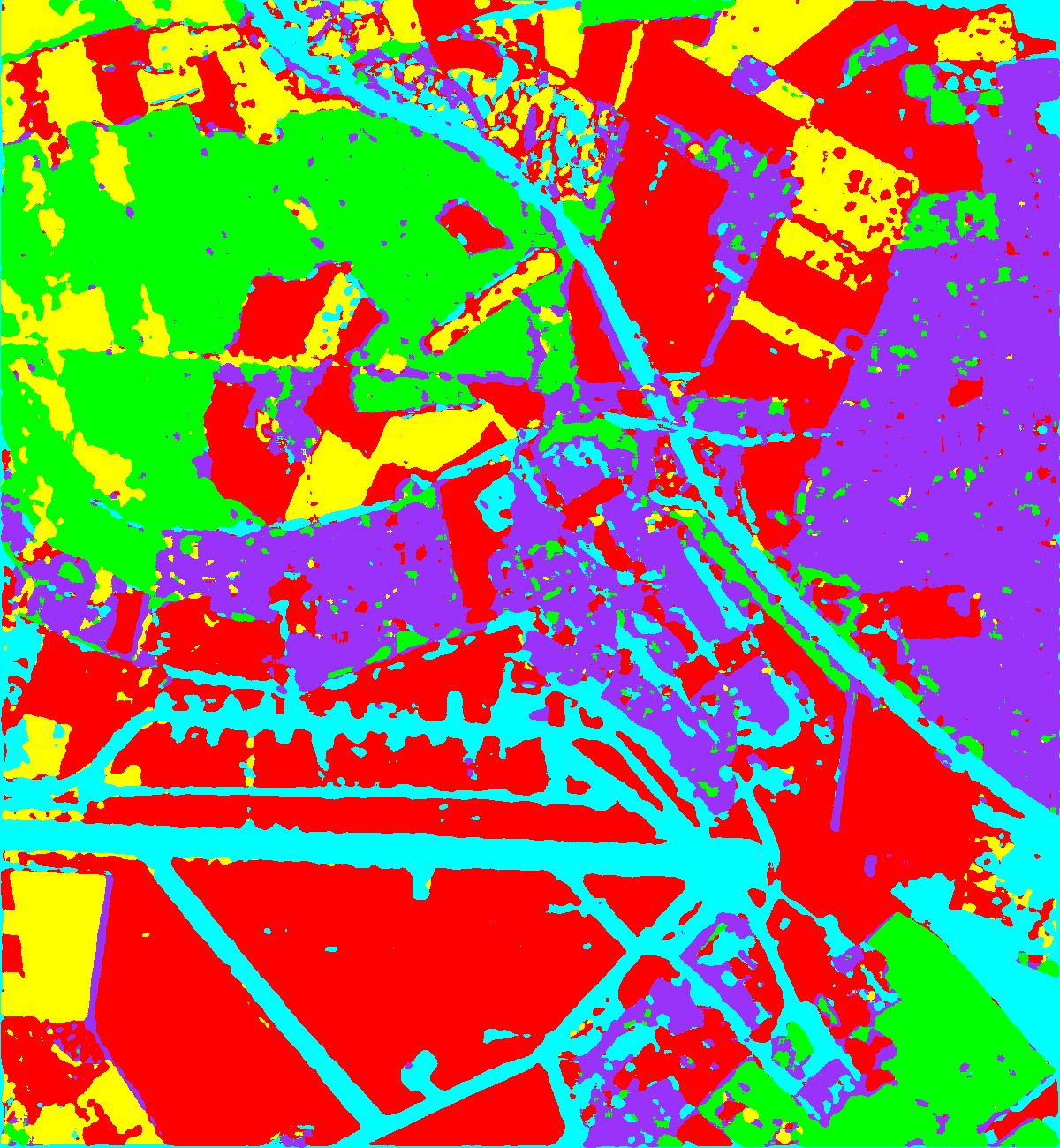}}}

\subfloat[]{\fbox{\includegraphics[height=0.14\textheight]{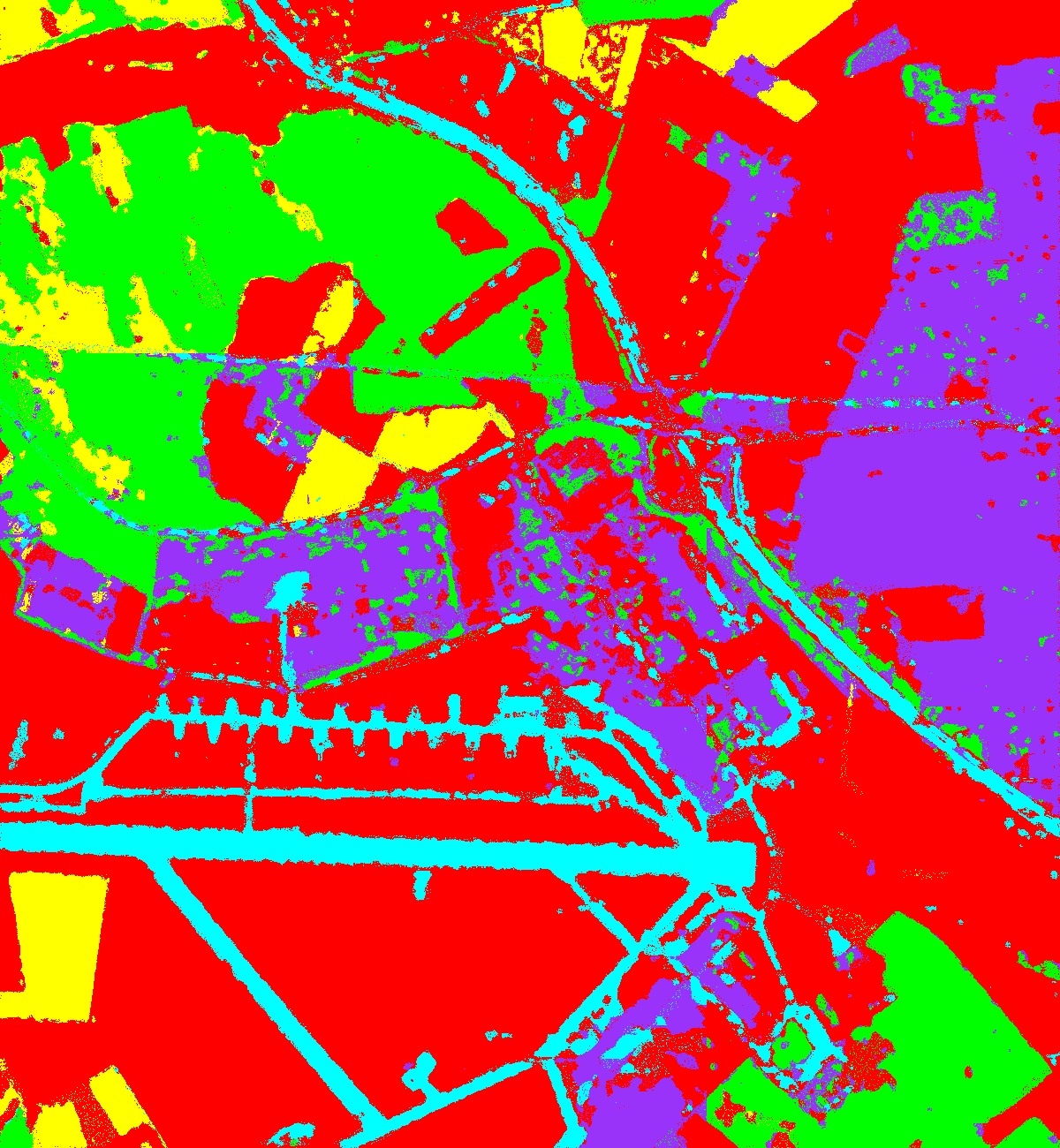}}}
\subfloat[]{\fbox{\includegraphics[height=0.14\textheight]{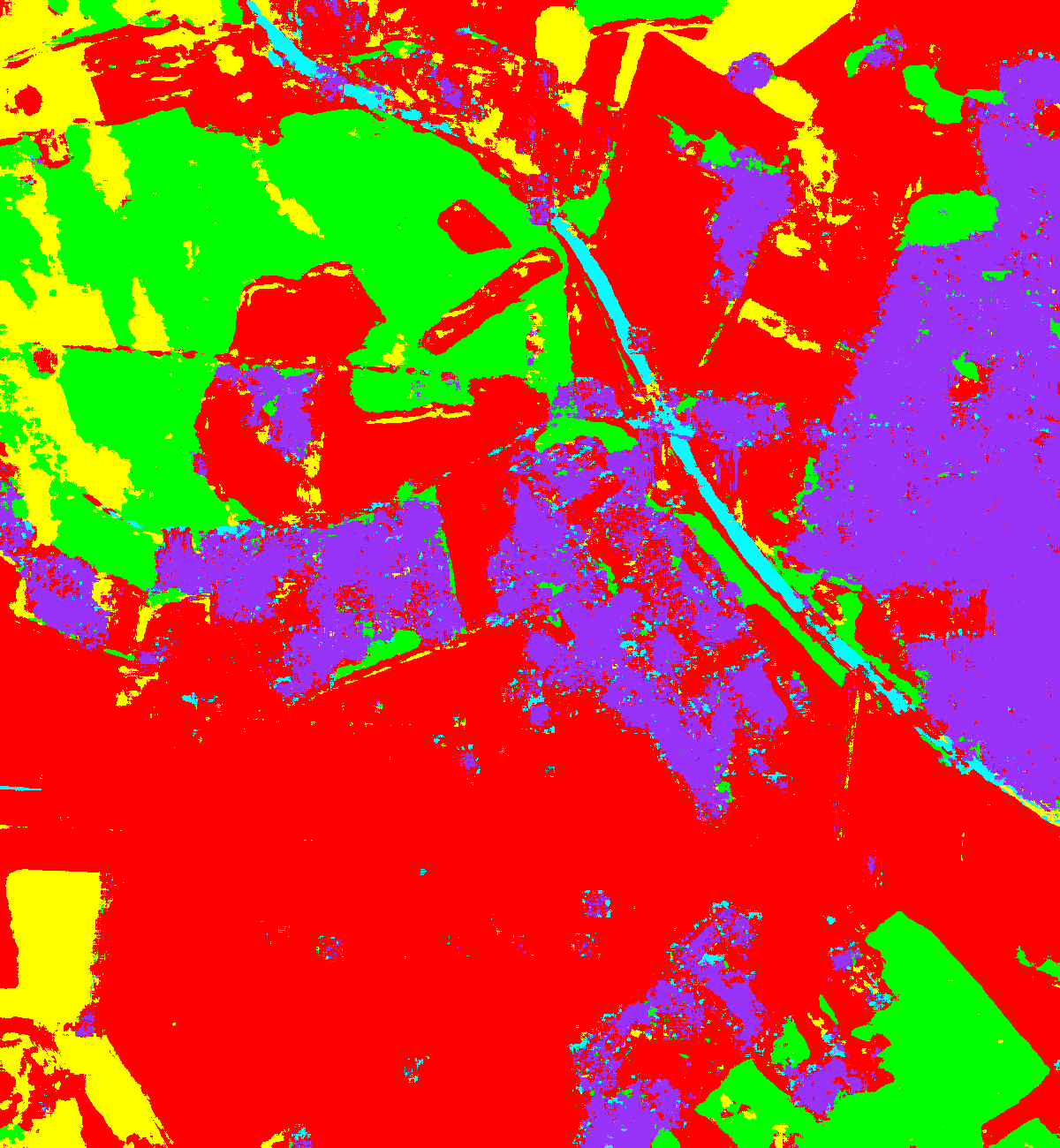}}}
\subfloat[]{\fbox{\includegraphics[height=0.14\textheight]{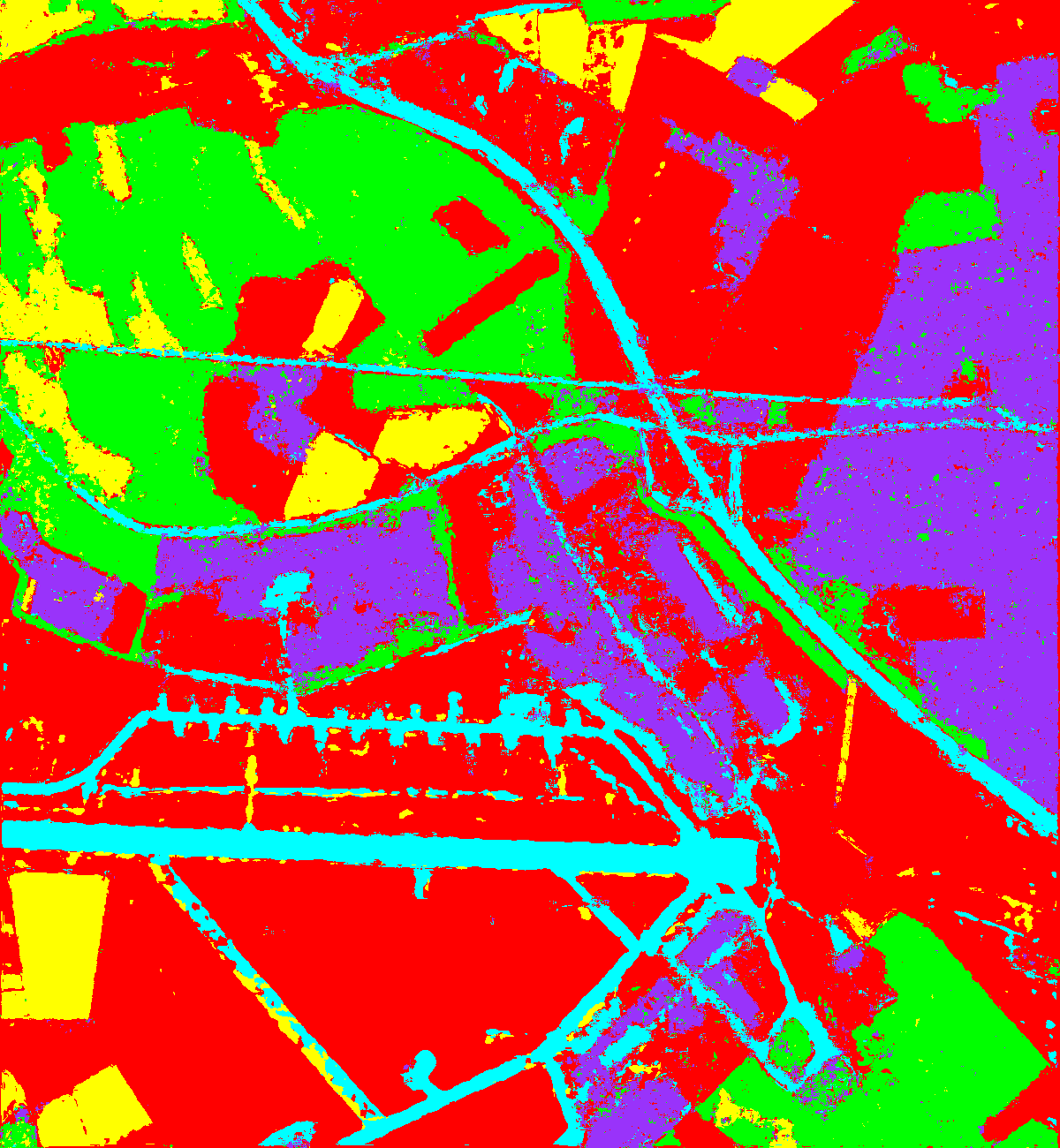}}}
\subfloat[]{\fbox{\includegraphics[height=0.14\textheight]{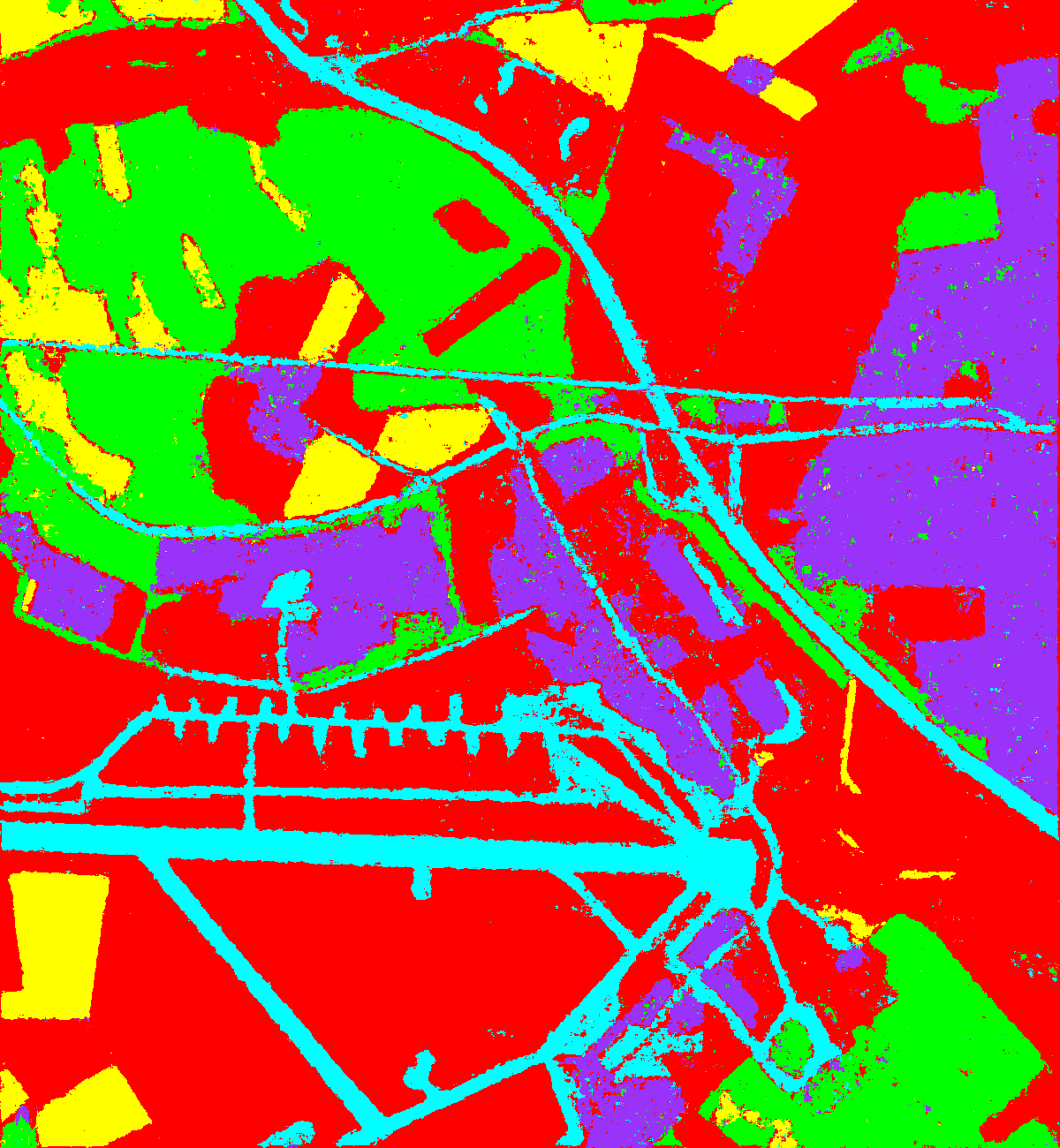}}}

\includegraphics[height=0.04\textheight]{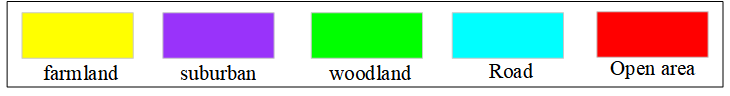}

\caption{Classification results of the Oberpfaffenhofen data set. (a) PauliRGB image of Oberpfaffenhofen area; (b) The label map of (a); (c) The classification map by the Super\_RF method; (d) The classification map by the CV-CNN method; (e) The classification map by the DFGCN method; (f) The classification map by the MPCNN method; (g) The classification map by the 9D-CNN method; (h) The classification map by the proposed RCM\_CNN method.}
\label{fig2}
\end{figure*}

Fig.\ref{fig2} gives the visual classification results generated by compared and proposed methods on Oberpfaffenhofen data set. Among them, experimental results by the four compared and proposed methods are illustrated in Figs.\ref{fig2} (c)-(f) and (h). In addition, the 9D-CNN method, as shown in Fig.\ref{fig2}(g), can be considered as the ablation study to evaluate the effectiveness of the proposed RCMnet module, which is different from the proposed RCM\_CNN method by replacing the RCM module by the CNN convolution layer. We can observe that the proposed method in (h) can obtain better classification performance than other methods in both heterogenous regions and edge details. The Super\_RF method in (c) produces some misclassifications in \emph{suburban} class and loses some edges in \emph{road}. The CVCNN in (d) cannot classify heterogenous regions well and loses some edge details, since it loses the channel correlation of covariance matrix by vectoring the matrix. The DFGCN in (e) can cause some misclassification in \emph{suburban} area, since it is a pixel-wise classification. The MPCNN in (f) performs well in region homogeneity of heterogenous terrain objects, while it almost totally loses the \emph{road}. The 9D-CNN in (g) produces many noisy classes. The proposed method reduces the noisy classes and improves the classification accuracy than 9D-CNN, which illustrates the effectiveness of the features learned by the RCMnet module.

\begin{figure}
\centering
\setlength{\fboxrule}{0.2pt}
  \setlength{\fboxsep}{0.01mm}
\includegraphics[height=0.2\textheight]{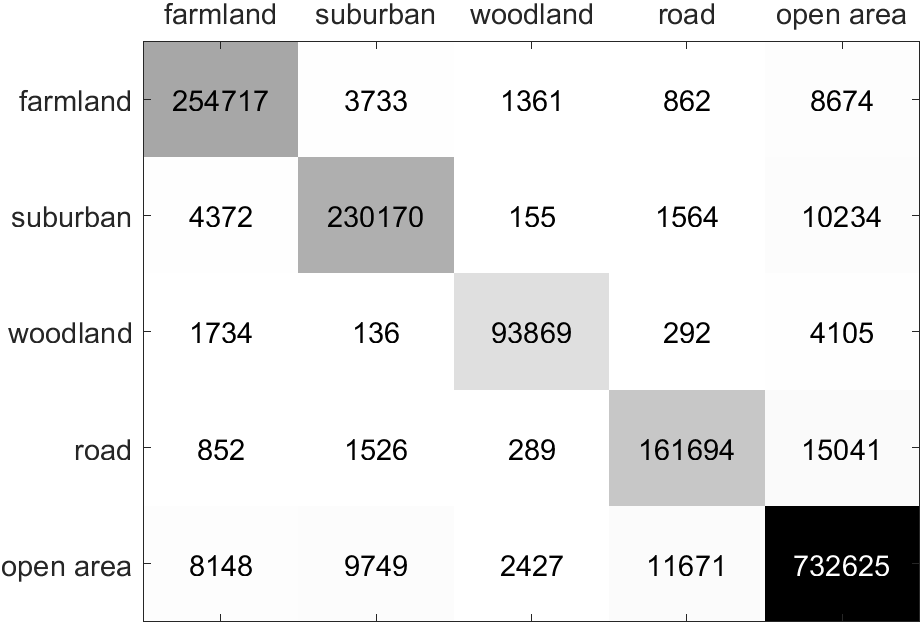}
\caption{Confusion matrix of the proposed RCM\_CNN method on Oberpfaffenhofen Data Set.}
\label{fig3}
\end{figure}

To evaluate the effectiveness of the proposed method quantitatively, the classification accuracy of different methods on Oberpfaffenhofen data set is presented in Table \ref{t2}. It can be seen that our method increases accuracies by 10.09\%, 19.55\%,  9.00\% and 17.61\% than other four compared methods respectively. In addition, it improves 4.34\% than 9D-CNN method, since it replaces the convolution layer by the RCM module, which demonstrates the effectiveness of the proposed RCMnet module. In addition, most compared methods cannot obtain markable performance in this PolSAR data set, since it is a challenging data set with complex terrain types, especially for the \emph{road}, \emph{suburban} and \emph{woodland} classes. The \emph{road} is difficult to be distinguished, and PolSAR data have sharp variations within the \emph{woodland} and \emph{suburban} area. Many state-of-the-art methods have failed on this PolSAR data set. However, our method can still achieve remarkable classification result. To be specific, the Super\_RF method cannot classify the \emph{road} well. The CVCNN method has low accuracies in both \emph{woodland} and \emph{open area} classes. The DFGCN also loses some \emph{road} areas. The MPCNN almost cannot tell out the \emph{road}. The 9D-CNN method cannot classify the \emph{road} well. The proposed method obtains the highest classification accuracies in various classes, and also achieves superior performance in OA, AA and Kappa coefficient.
Furthermore, the confusion matrix of the proposed method is illustrated in Fig.\ref{fig3}. It is found that amount of pixels in \emph{suburban} and \emph{road} classes are misclassified as \emph{open area}.  The \emph{open area} has higher classification accuracy than other classes.

\begin{table*}[ht]
\footnotesize
\begin{center}
\caption
{ \label{t2}
 Classification accuracy of different methods on Oberpfaffenhofen Data Set (\%).}
\begin{tabular}{p{1.8cm}p{1.4cm}p{1.3cm}p{1.3cm}p{1.3cm}p{1.3cm}p{1.3cm}p{2.0cm}}
\hline
class&Super\_RF&CVCCN&DFGCN&MPCNN&9D-CNN&RCM\_CNN\\
\hline
farmland&91.64&81.16&85.08&82.51&93.00&\textbf{94.57}\\
suburban&86.95&87.52&87.30&85.82&91.82&\textbf{93.38}\\
woodland&83.60&70.49&77.57&65.65&89.63&\textbf{93.74}\\
road&64.07&76.20&66.10&10.38&73.78&\textbf{90.12}\\
open area&61.68&68.86&90.51&86.92&92.39&\textbf{95.82}\\
OA&84.34&74.88&85.43&75.82&90.09&\textbf{94.43}\\
AA&77.59&76.85&81.31&66.26&88.12&\textbf{93.53}\\
Kappa&76.67&65.73&78.66&63.49&85.56&\textbf{91.88}\\
\hline
\end{tabular}
\end{center}
\end{table*}

\subsection{Experimental results on Xi'an data set}

\begin{figure*}
\centering
\setlength{\fboxrule}{0.2pt}
  \setlength{\fboxsep}{0.01mm}
\subfloat[]{\includegraphics[height=0.14\textheight]{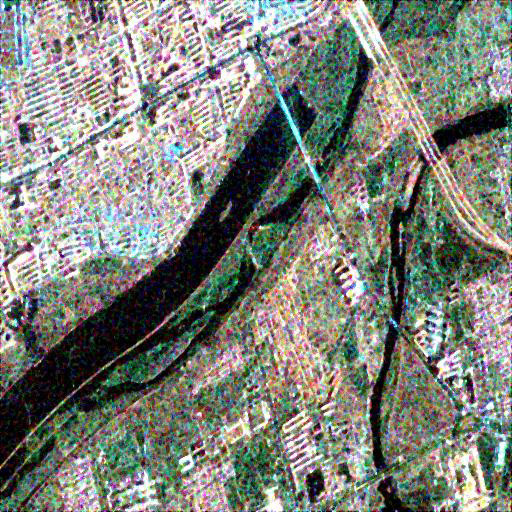}}
\subfloat[]{\fbox{\includegraphics[height=0.14\textheight]{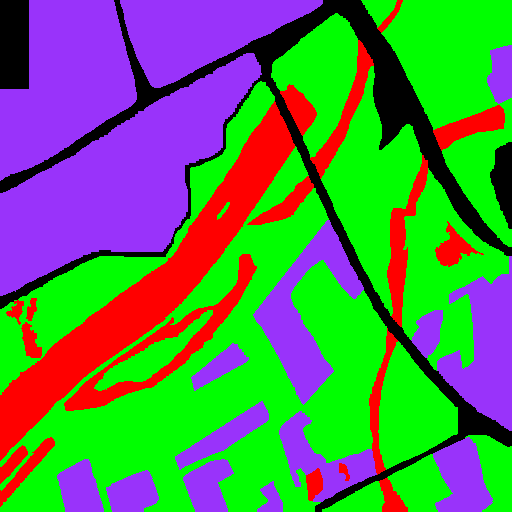}}}
\subfloat[]{\fbox{\includegraphics[height=0.14\textheight]{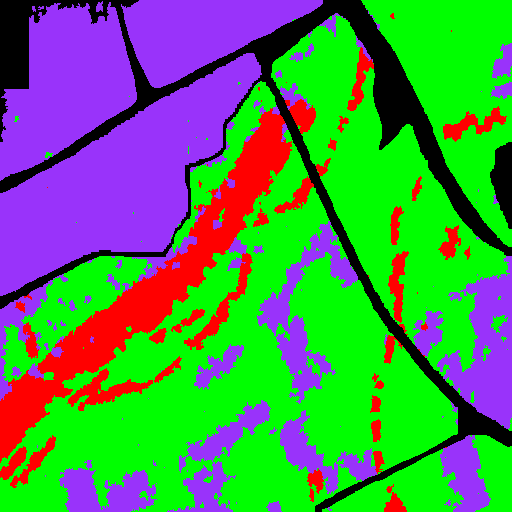}}}
\subfloat[]{\fbox{\includegraphics[height=0.14\textheight]{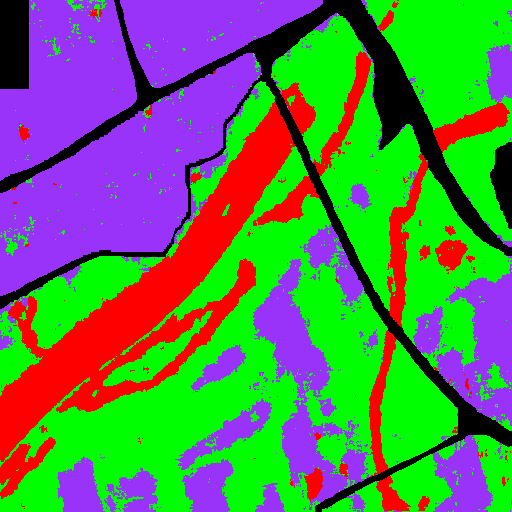}}}

\subfloat[]{\fbox{\includegraphics[height=0.14\textheight]{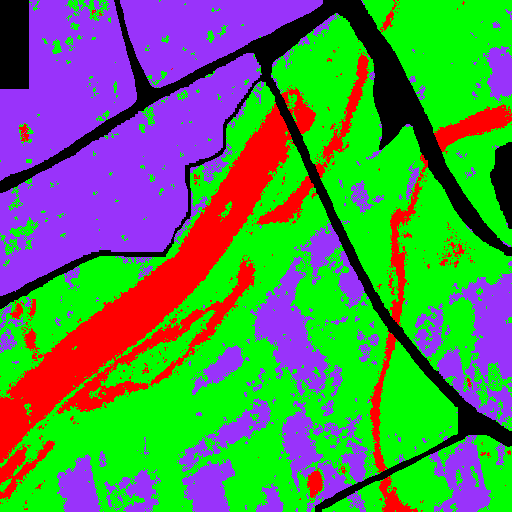}}}
\subfloat[]{\fbox{\includegraphics[height=0.14\textheight]{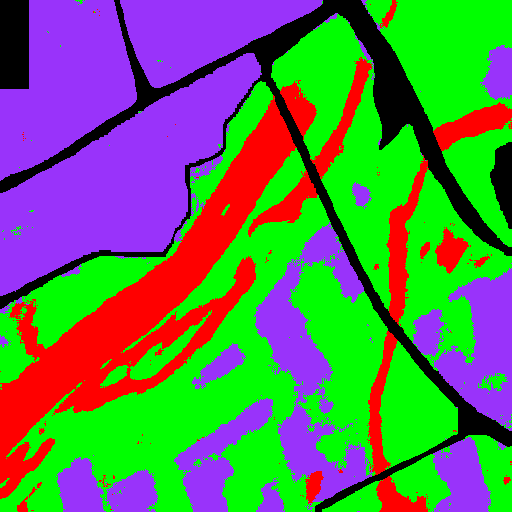}}}
\subfloat[]{\fbox{\includegraphics[height=0.14\textheight]{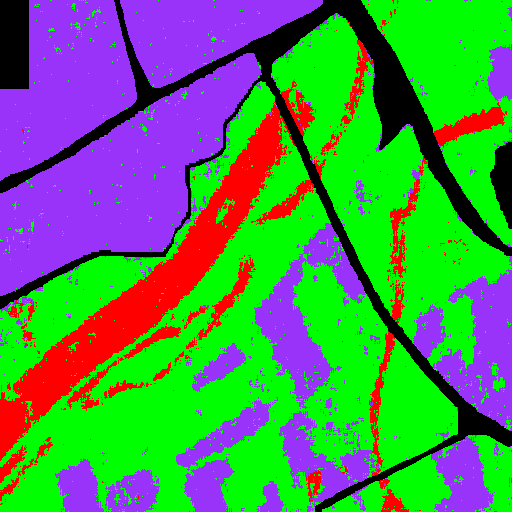}}}
\subfloat[]{\fbox{\includegraphics[height=0.14\textheight]{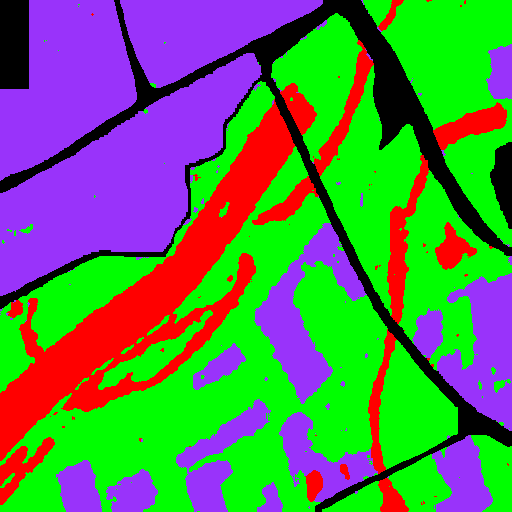}}}

\includegraphics[height=0.04\textheight]{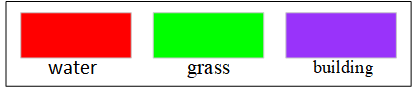}

\caption{Classification results of the  Xi'an data set. (a) PauliRGB image of Xi'an area; (b) The label map of (a); (c) The classification map by the Super\_RF method; (d) The classification map by the CV-CNN method; (e) The classification map by the DFGCN method; (f) The classification map by the MPCNN method; (g) The classification map by the 9D-CNN method; (h) The classification map by the proposed RCM\_CNN method.}
\label{fig4}
\end{figure*}

The visual classification maps by different methods are given in Fig.\ref{fig4}, in which Figs.\ref{fig4}(c)-(g) are the classification results by the Super\_RF, CVCNN, DFGCN, MPCNN and 9D-CNN methods respectively. The classification map by the proposed method is presented in Fig.\ref{fig4}(h). It can be seen that the proposed method can achieve obvious classification advantage compared with other methods in both \emph{building} and \emph{water} areas. Specifically, the Super\_RF method loses many \emph{water} areas in (c). The CVCNN causes some misclassifications in heterogenous regions and loses many edge details in (d). The DFGCN cannot classify \emph{buildings} well in (e). The  MPGCN can improve the classification performance. The 9D-CNN method causes some mistakes especially in \emph{water} area. The proposed method can obtain superior performance in both \emph{water} and \emph{building} areas. In particular, compared with 9D-CNN, the proposed method can obtain more accurate classification, which demonstrates the effectiveness of the feature learned by RCMnet module. Therefore, Remmianan complex matrix network is necessary for PolSAR image classification.

\begin{figure}
\centering
\setlength{\fboxrule}{0.2pt}
  \setlength{\fboxsep}{0.01mm}
\includegraphics[height=0.2\textheight]{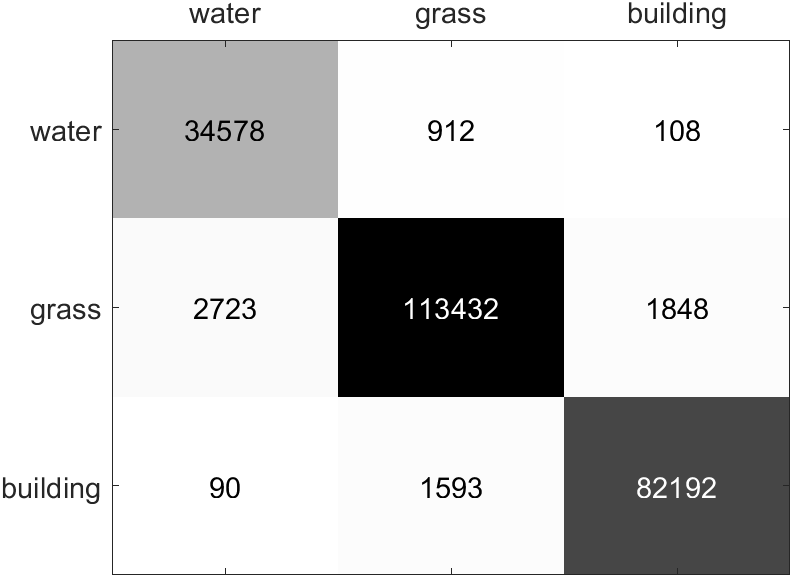}
\caption{Confusion matrix of the proposed RCM\_CNN method on Xi'an Data Set.}
\label{fig5}
\end{figure}

In addition, the classification accuracy and some evaluation metrics are calculated to quantitatively evaluate the effectiveness of the proposed method in Table \ref{t3}. From Table \ref{t3}, we can see that our method improve the classification accuracy greatly than other methods by 7.00\%, 4.57\%, 7.4\%, 2.93\% respectively. Specifically, the Super\_RF method has lower classification accuracy in \emph{water} class. The CVCNN method has low accuracies in both \emph{woodland} and \emph{open area} classes. The DFGCN cannot classify \emph{water} well. The MPCNN cannot classify the \emph{grass} well. The 9D-CNN method cannot classify the \emph{water} well. The proposed method can obtain the highest classification accuracies in various classes, and also achieve superior performance in OA, AA and Kappa coefficient.
In addition, as the ablation study, the 9D-CNN method is 2.79\% lower than our method, which verifies the features learned from RCM module in Riemannian space is better than CNN features in Euclidian space.
Moreover, Fig.\ref{fig5} presents the confusion matrix of the proposed method. From Fig.\ref{fig5} we can see that the main confusion is \emph{grass} and \emph{water} classes, which is consistent with the visual classification map in Fig.\ref{fig4}(h). Some pixels in \emph{grass} area are misclassified as \emph{water} class.

\begin{table*}[ht]
\footnotesize
\begin{center}
\caption
{ \label{t3}
 Classification accuracy of different methods on Xi'an Data Set (\%).}
\begin{tabular}{p{1.8cm}p{1.4cm}p{1.3cm}p{1.3cm}p{1.3cm}p{1.3cm}p{1.3cm}p{2.0cm}}
\hline
class&Super\_RF&CVCCN&DFGCN&MPCNN&9D-CNN&RCM\_CNN\\
\hline
water&70.91&94.55&86.15&95.52&91.94&\textbf{97.13}\\
grass&90.94&90.68&88.36&90.95&93.52&\textbf{96.13}\\
city&94.97&93.81&92.65&97.68&95.97&\textbf{97.99}\\
OA&89.94&92.37&89.54&94.01&94.15&\textbf{96.94}\\
AA&85.61&93.01&89.05&94.71&93.81&\textbf{97.08}\\
Kappa&83.02&87.51&82.75&90.25&90.36&\textbf{94.97}\\
\hline
\end{tabular}
\end{center}
\end{table*}

\subsection{Experimental results on San Francisco data set}

\begin{figure*}
\centering
\setlength{\fboxrule}{0.2pt}
  \setlength{\fboxsep}{0.01mm}
\subfloat[]{\includegraphics[height=0.18\textheight]{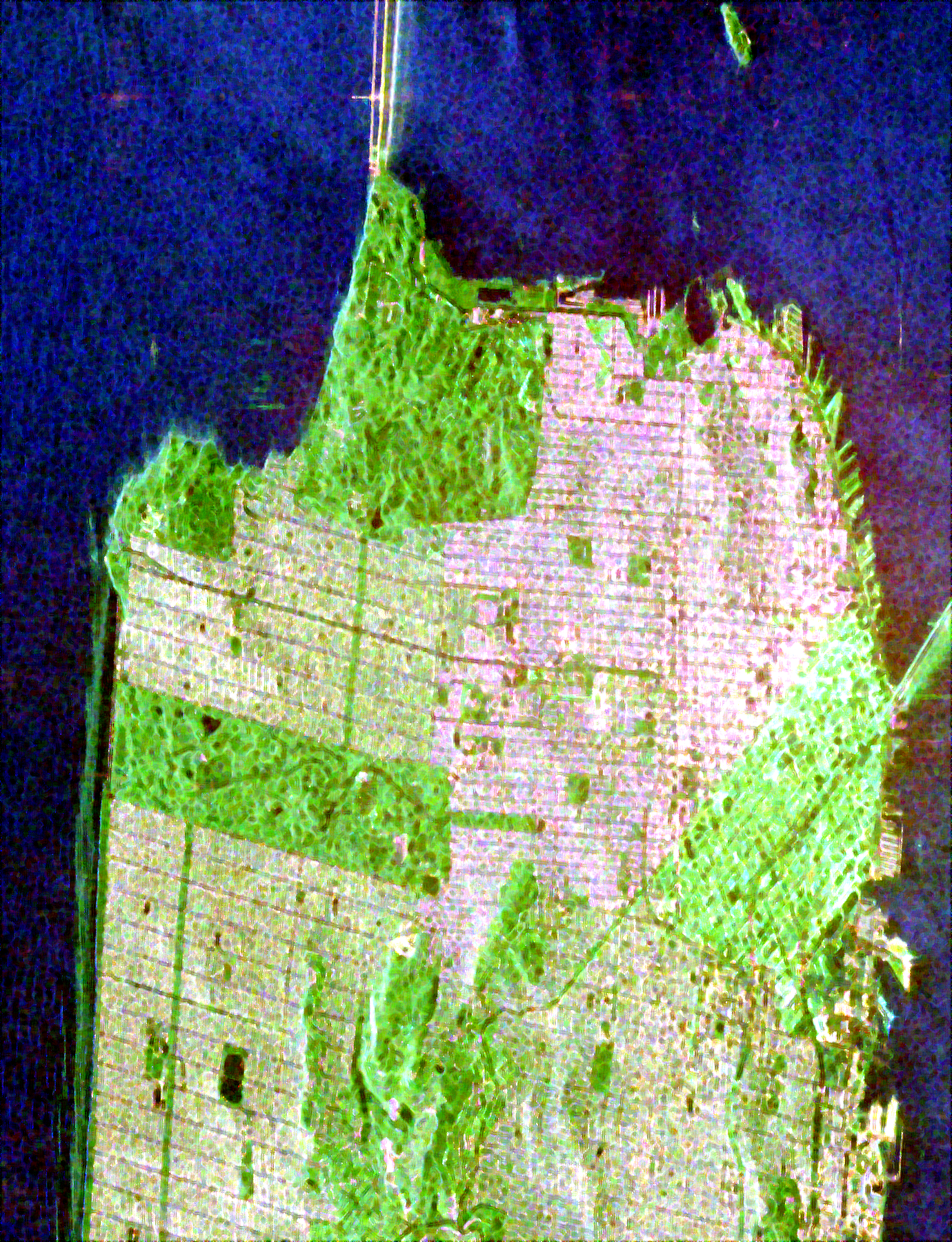}}
\subfloat[]{\fbox{\includegraphics[height=0.18\textheight]{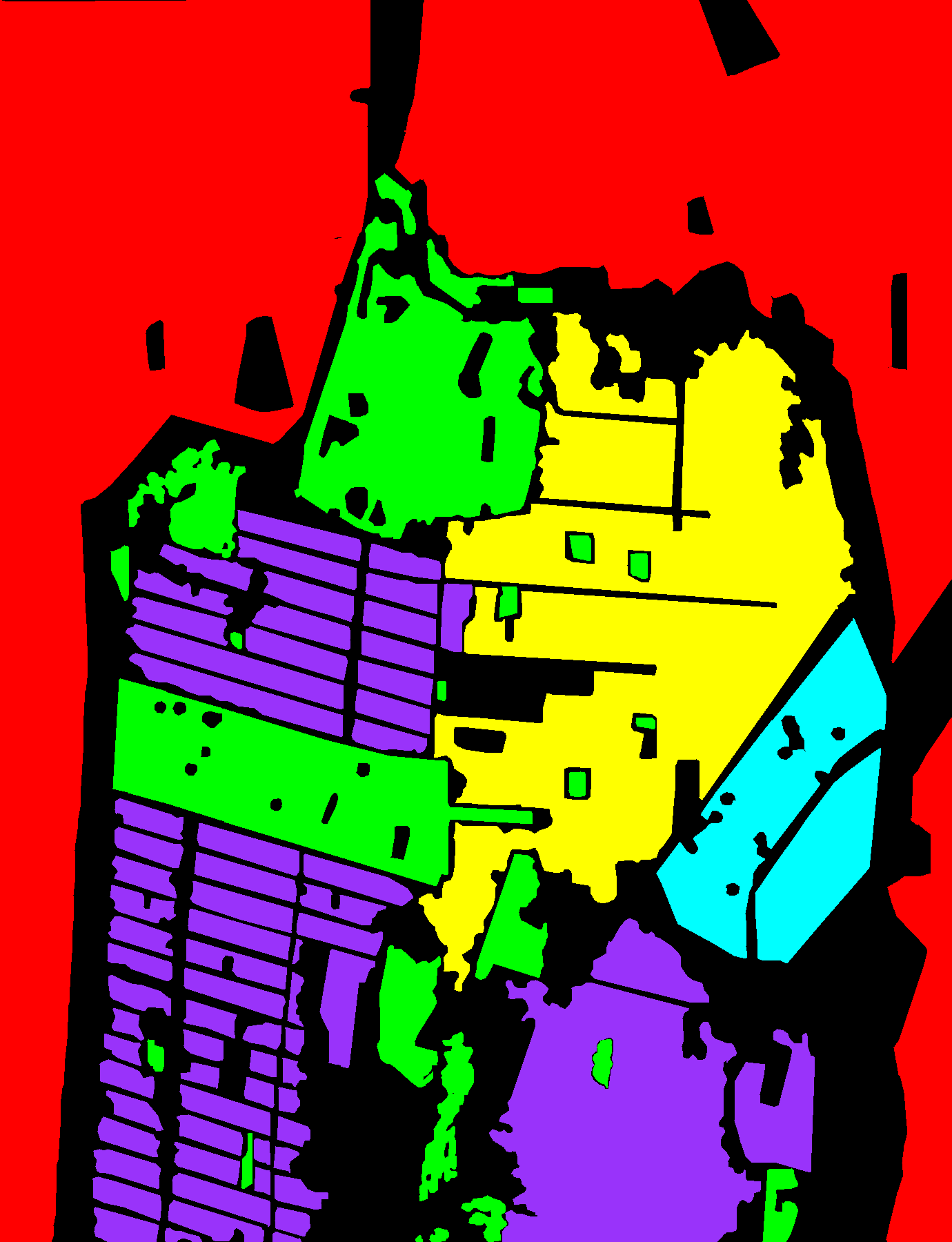}}}
\subfloat[]{\fbox{\includegraphics[height=0.18\textheight]{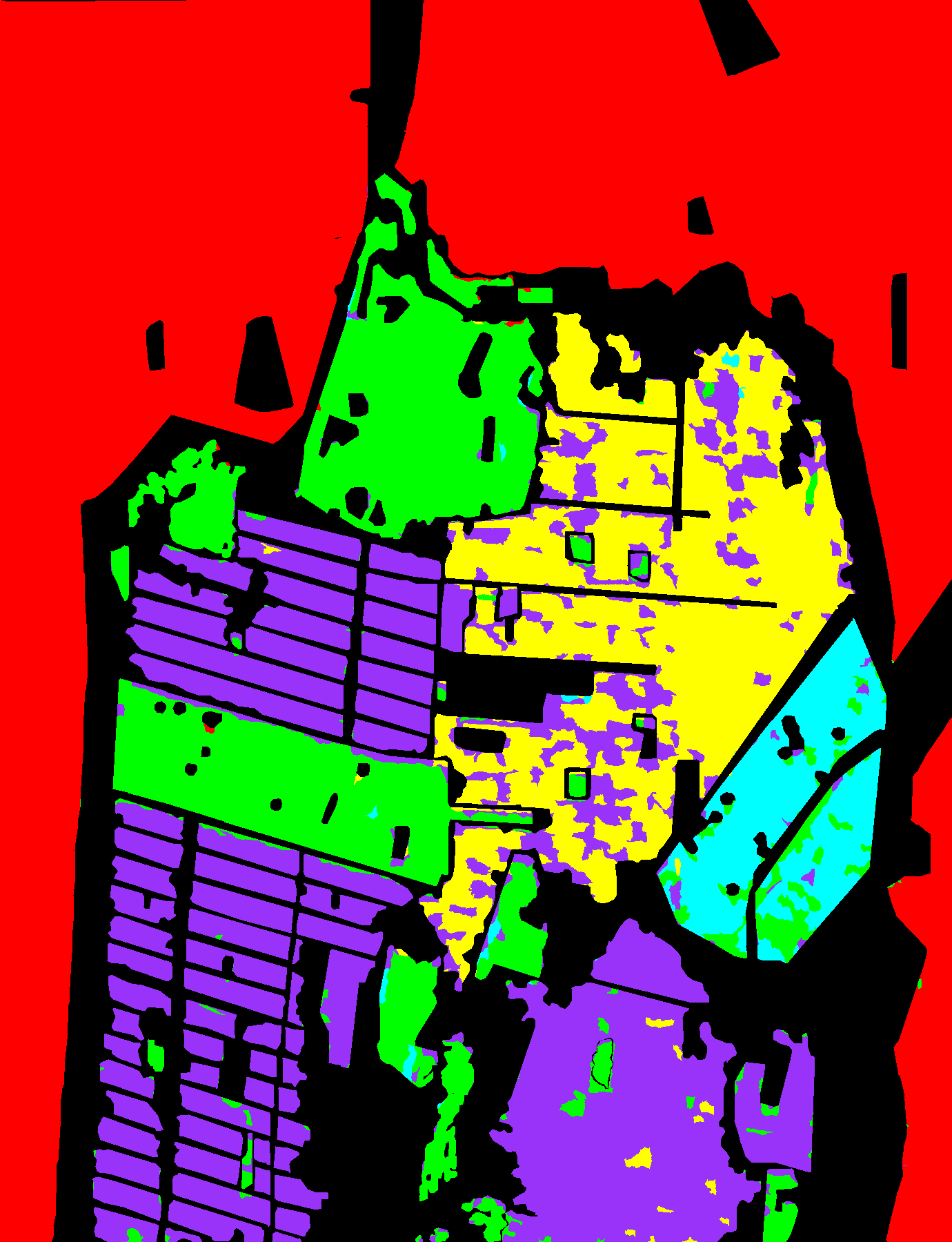}}}
\subfloat[]{\fbox{\includegraphics[height=0.18\textheight]{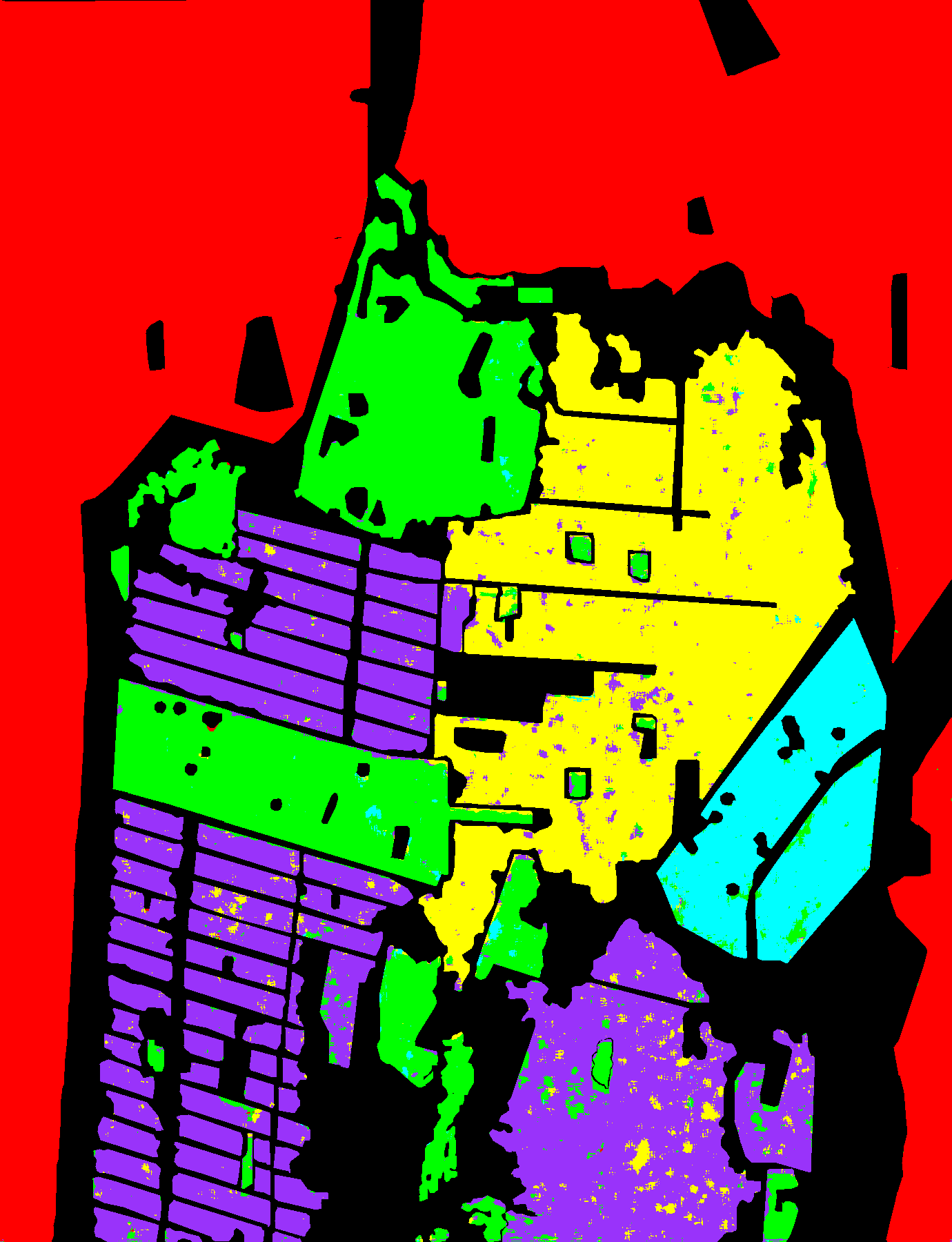}}}

\subfloat[]{\fbox{\includegraphics[height=0.18\textheight]{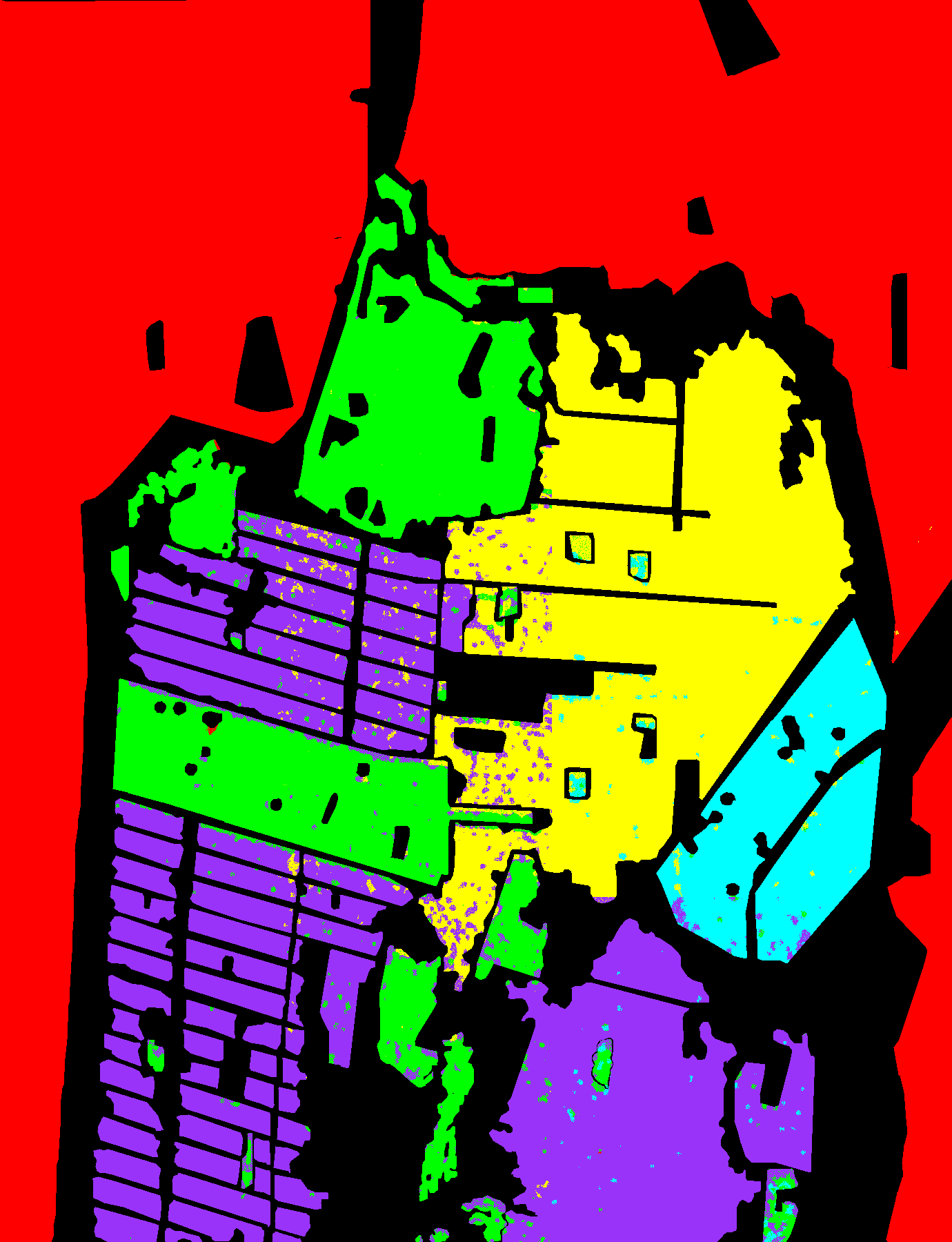}}}
\subfloat[]{\fbox{\includegraphics[height=0.18\textheight]{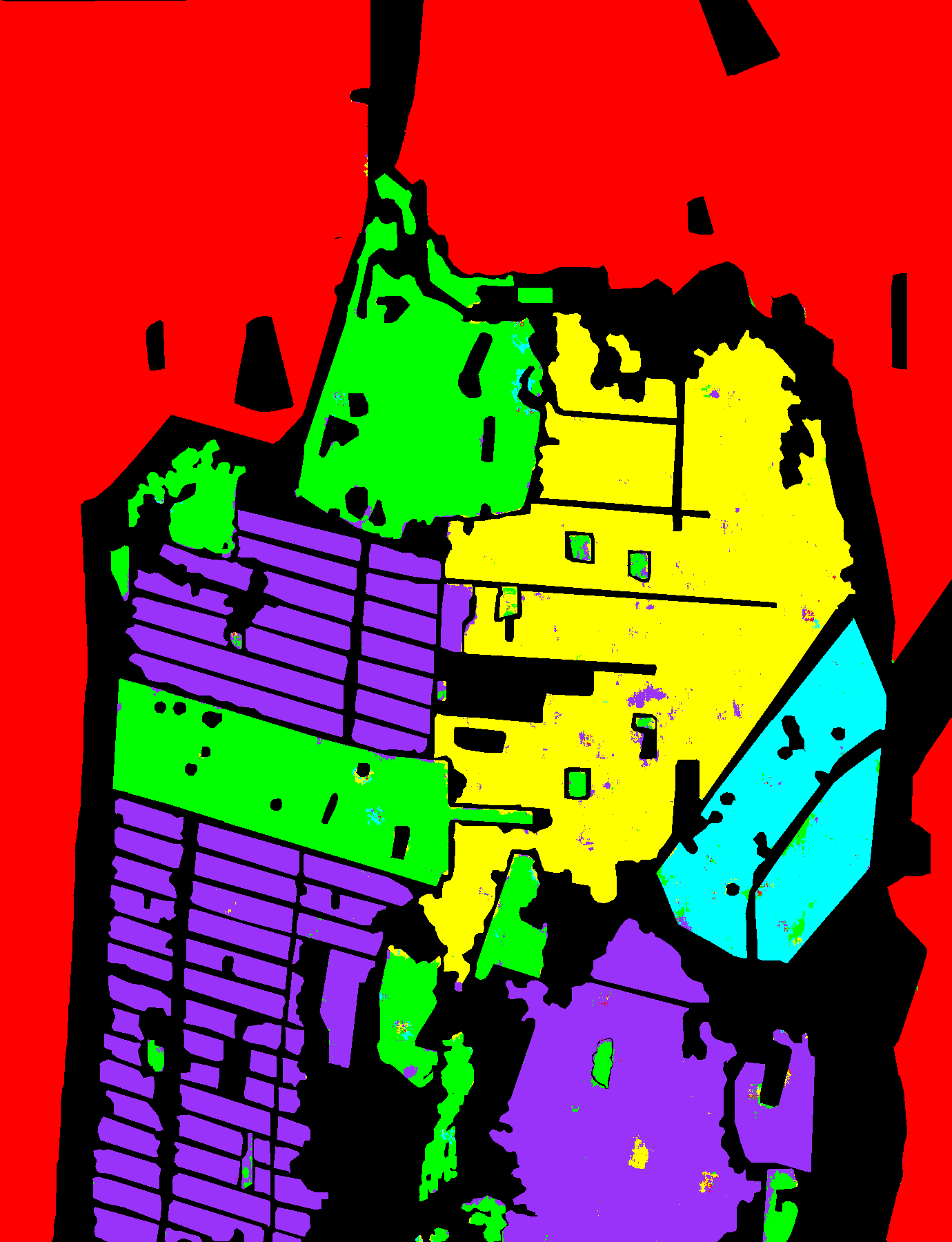}}}
\subfloat[]{\fbox{\includegraphics[height=0.18\textheight]{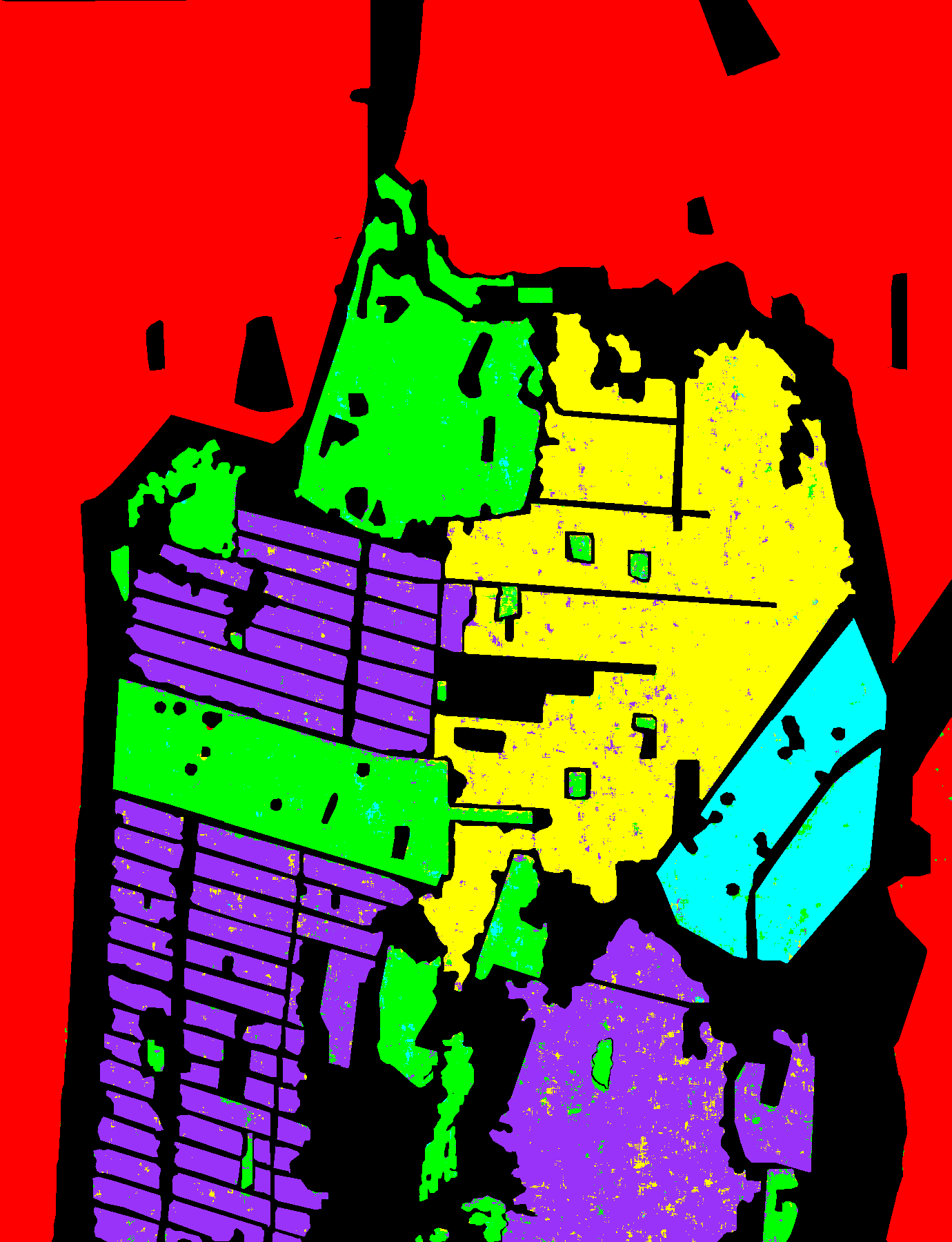}}}
\subfloat[]{\fbox{\includegraphics[height=0.18\textheight]{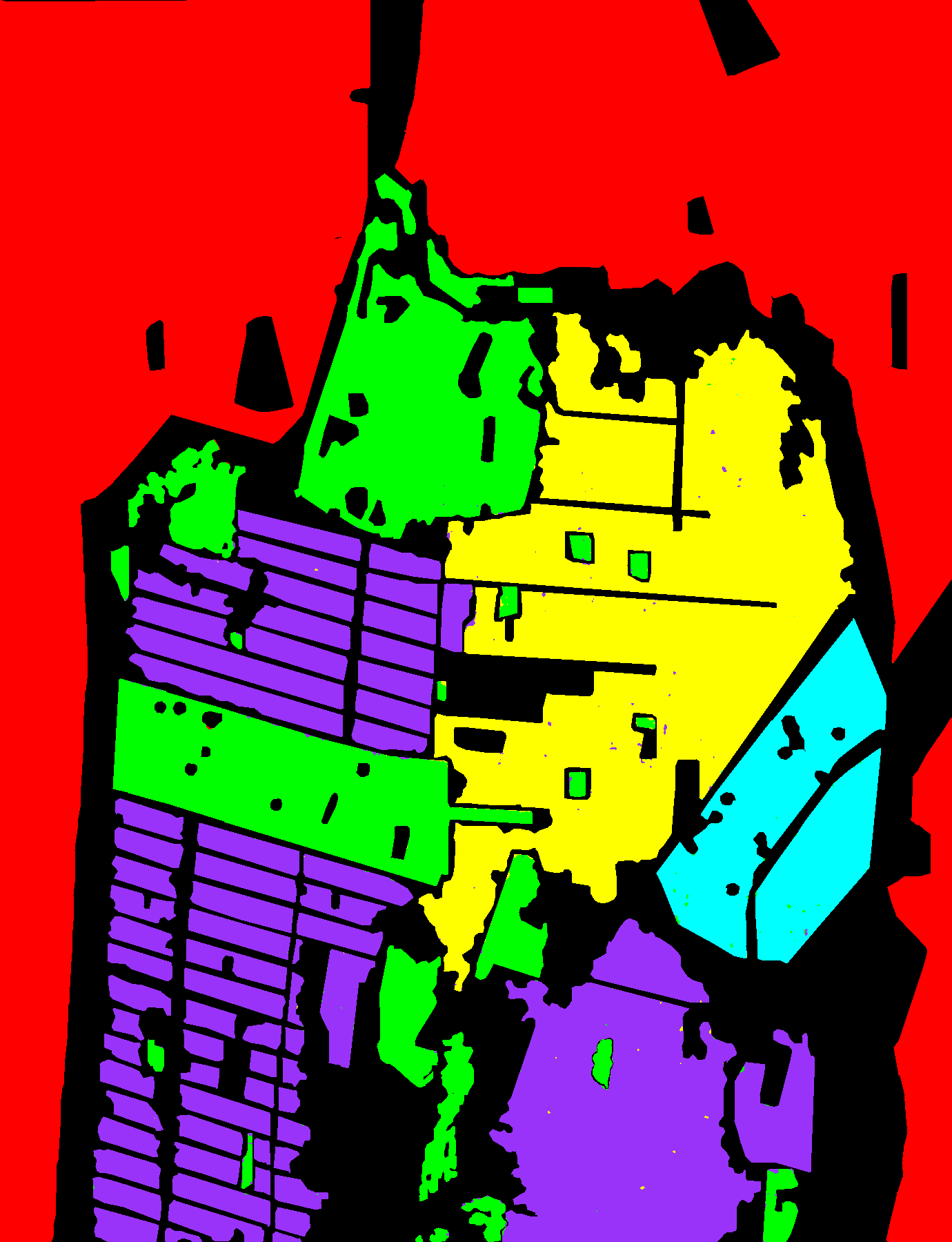}}}

\includegraphics[height=0.04\textheight]{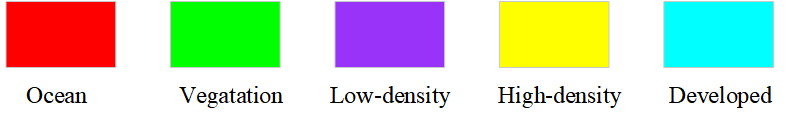}

\caption{Classification results of the San Francisco data set. (a) PauliRGB image of San Francisco area; (b) The label map of (a); (c) The classification map by the Super\_RF method; (d) The classification map by the CV-CNN method; (e) The classification map by the DFGCN method; (f) The classification map by the MPCNN method; (g) The classification map by the 9D-CNN method; (h) The classification map by the proposed RCM\_CNN method.}
\label{fig6}
\end{figure*}

The visual classification maps by four compared and proposed methods are presented in Figs.\ref{fig6}(c)-(f) and (h) respectively. In addition, the ablation study in (g) is the classification map by the 9D-CNN method, which is given to verify the effectiveness of the proposed RCMnet. It can be seen that many methods can achieve sound performance in this PolSAR data set, and the proposed method produces superior performance. To be specific, the Super\_RF method in (c) cannot classify the \emph{high-density} area well, since it is a shallow level feature learning method without deep learning. The CVCNN method in (d) causes many mistakes in \emph{high-density} and \emph{low-density} urban areas. The DFGCN method also produces some misclassifications in urban area in (e). The MPCNN method in (f) has some noisy classes in \emph{high-density} urban area. The 9D-CNN method still produces some noisy classes in both \emph{low-density} and \emph{high-density} urban areas. Compared with other methods, the proposed method can obtain superior classification performance in both heterogenous and homogenous regions. In addition, the proposed method can also obtain better classification result than the 9D-CNN method for this PolSAR image, which once again proves the effectiveness of the proposed RCMnet module.

\begin{figure}
\centering
\setlength{\fboxrule}{0.2pt}
  \setlength{\fboxsep}{0.01mm}
\includegraphics[height=0.2\textheight]{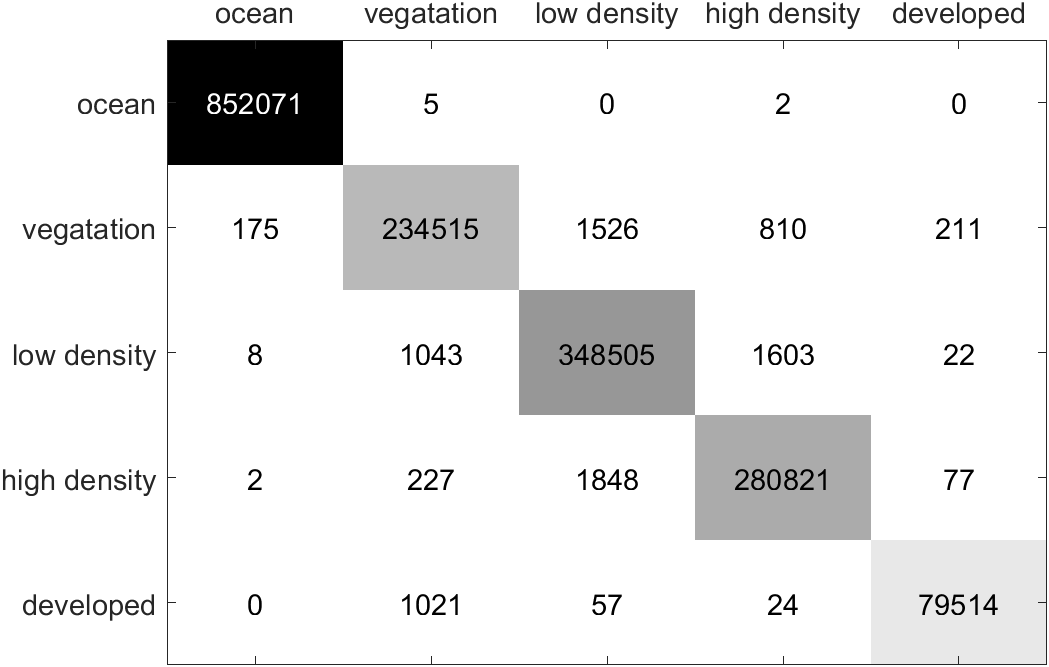}
\caption{Confusion matrix of the proposed RCM\_CNN method on San Francisco Data Set.}
\label{fig7}
\end{figure}

Moreover, the classification accuracies of compared and proposed methods are illustrated in Table \ref{t4}. From Table \ref{t4}, we can see that our method increases the classification accuracy than other compared methods by 5.19\%, 1.81\%, 1.62\%, 0.59\% respectively. To be specific, the Super\_RF method has low accuracy in \emph{high-density} class. The CVCNN method cannot classify the \emph{low-density} well. Both the DFGCN and MPCNN methods have low accuracies in \emph{vegetation} class. Besides, the difference of our method from 9D-CNN is the utilization of the RCM module rather than a CNN convolution layer. It can improve 1.16\% by using RCM module. Overall, the proposed method can obtain the highest classification accuracies in various classes, and also achieve superior performance in OA, AA and Kappa coefficient.
Furthermore,the confusion matrix of the proposed method is given in Fig.\ref{fig7}. It can seen that the main confusion is \emph{low-density} and \emph{high-density} classes.

\begin{table*}[ht]
\footnotesize
\begin{center}
\caption
{ \label{t4}
 Classification accuracy of different methods on San Francisco Data Set (\%).}
\begin{tabular}{p{1.8cm}p{1.4cm}p{1.3cm}p{1.3cm}p{1.3cm}p{1.3cm}p{1.3cm}p{2.0cm}}
\hline
class&Super\_RF&CVCCN&DFGCN&MPCNN&9D-CNN&RCM\_CNN\\
\hline
ocean&99.98&99.99&\textbf{100}&99.98&99.98&\textbf{100}\\
vegetation&93.89&96.42&93.74&95.63&95.55&\textbf{98.85}\\
low-density&97.31&94.51&97.26&99.39&96.46&\textbf{99.24}\\
high-density&77.76&96.37&96.17&98.57&97.60&\textbf{99.24}\\
developed&81.00&95.92&96.97&96.81&97.52&\textbf{98.63}\\
OA&94.33&97.71&97.90&98.93&98.36&\textbf{99.52}\\
AA&89.99&96.64&96.83&98.08&97.62&\textbf{99.19}\\
Kappa&91.81&96.70&96.98&98.46&97.64&\textbf{99.31}\\
\hline
\end{tabular}
\end{center}
\end{table*}

\subsection{Discussion}

\emph{1)Effect of the layer number of the RCM backbone}

To analyze the layer number of the proposed RCM module on classification performance, we design different layers of RCM backbone, which are conducted on the three PolSAR data sets respectively. They are RCM0\_CNN, RCM1\_CNN, RCM2\_CNN methods respectively. The RCM0\_CNN method is the 9D-CNN method without RCM backbone, which replaces the RCM module by a CNN convolution layer. The experimental results have been shown in Figs.\ref{fig2}, \ref{fig4} and \ref{fig6}(g). The RCM1\_CNN method utilizes one layer RCM backbone followed by the CNN model. The RCM2\_CNN method uses two layer RCM backbone, and on the tail with the CNN model. The RCM0\_CNN, RCM1\_CNN and RCM2\_CNN can analyze the effect of the layer number of RCM backbone on classification performance.

The classification accuracies of different layer number of RCM module on Oberpfaffenhofen, Xi'an and San Francisco data sets are given in Table \ref{t5}, \ref{t6} and \ref{t7} respectively. It can be seen that RCM1\_CNN has obvious accuracy improvements than RCM0\_CNN method on all the three types of PolSAR images. It is because RCMnet can learn the geometric features of complex matrix in Riemannian space, which can convert the complex matrix into a low-dimension separable feature space. Specifically, the RCM1\_CNN method can improve by 4.34\%, 2.79\% and 1.16\% in OA than the RCM0\_CNN method respectively on Oberpfaffenhofen, Xi'an and San Francisco data sets. However, it also can be seen that the RCM2\_CNN method only improve the classification accuracy by 0.02\%,0.04\% and 0.03\% than RCM1\_CNN  method on the three sets of PolSAR data respectively. It shows that increasing the layer number cannot greatly improve the classification accuracy, and the single RCM layer is effective to improve the results. Therefore, we select one layer RCM module in the proposed method, which can greatly reduce the computation time either.

\begin{table}[ht]
\footnotesize
\begin{center}
\caption
{ \label{t5}
 Classification accuracy of different RCM layer number on Oberpfaffenhofen Data Set (\%).}
\begin{tabular}{p{1.6cm}p{1.4cm}p{1.4cm}p{1.4cm}}
\hline
&RCM0\_CNN&RCM1\_CNN&RCM2\_CNN\\
\hline
farmland&93.00&94.57&\textbf{94.80}\\
suburban&91.82&93.38&\textbf{94.36}\\
woodland&89.63&93.74&\textbf{94.23}\\
road&73.78&90.12&\textbf{91.71}\\
open area&92.39&\textbf{95.82}&94.96\\
OA&90.09&94.43&\textbf{94.45}\\
AA&88.12&93.53&\textbf{93.89}\\
Kappa&85.56&91.88&\textbf{91.95}\\
\hline
\end{tabular}
\end{center}
\end{table}

\begin{table}[ht]
\footnotesize
\begin{center}
\caption
{ \label{t6}
 Classification accuracy of different RCM layer number on Xi'an Data Set (\%).}
\begin{tabular}{p{1.6cm}p{1.4cm}p{1.4cm}p{1.4cm}}
\hline
&RCM0\_CNN&RCM1\_CNN&RCM2\_CNN\\
\hline
water&91.94&97.13&\textbf{97.15}\\
grass&93.52&96.13&\textbf{96.25}\\
buildings&95.97&\textbf{97.99}&97.79\\
OA&94.15&96.94&\textbf{96.98}\\
AA&93.81&97.08&\textbf{97.11}\\
Kappa&90.36&\textbf{94.97}&94.87\\
\hline
\end{tabular}
\end{center}
\end{table}

\begin{table}[ht]
\footnotesize
\begin{center}
\caption
{ \label{t7}
 Classification accuracy of different RCM layer number on San Francisco Data Set (\%).}
\begin{tabular}{p{1.6cm}p{1.4cm}p{1.4cm}p{1.4cm}}
\hline
&RCM0\_CNN&RCM1\_CNN&RCM2\_CNN\\
\hline
ocean&99.98&\textbf{100}&\textbf{100}\\
vegetation&95.55&98.85&\textbf{98.88}\\
low-density&96.46&99.24&\textbf{99.43}\\
high-density&97.60&\textbf{99.24}&99.16\\
developed&97.52&\textbf{98.63}&98.51\\
OA&98.36&99.52&\textbf{99.53}\\
AA&97.62&\textbf{99.19}&99.16\\
Kappa&97.64&99.31&\textbf{99.34}\\
\hline
\end{tabular}
\end{center}
\end{table}


\emph{2)Effect of the number of training samples}

\begin{figure}
\centering
\setlength{\fboxrule}{0.2pt}
  \setlength{\fboxsep}{0.01mm}
\includegraphics[height=0.2\textheight]{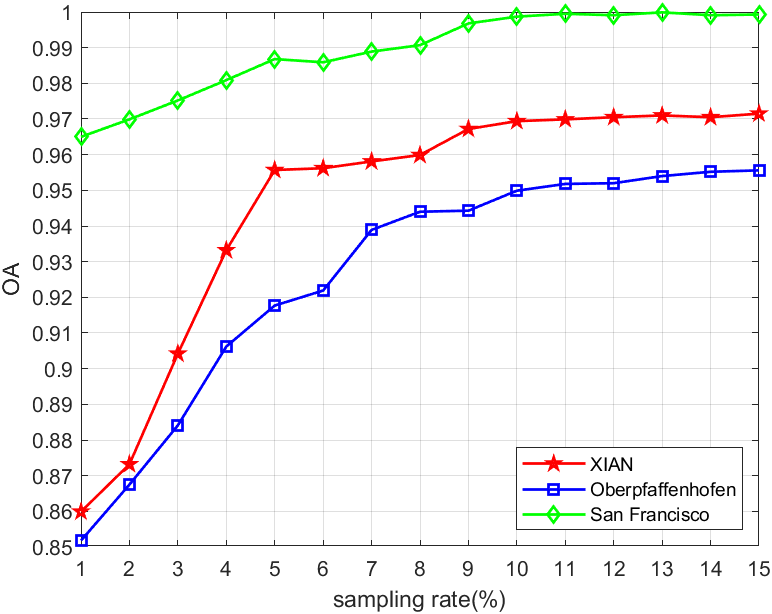}
\caption{The effect of the ratio of training samples on classification accuracy.}
\label{fig8}
\end{figure}

The number of training samples is an important parameter in the proposed method, which determines the performance of network learning. To test the effect of sampling rate on classification accuracy, experiments are conducted on the three sets of PolSAR  data above and different ratios of training samples are tested. The variation of classification accuracies with the ratio of training samples is given in Fig.\ref{fig8}. It can be seen that the classification accuracy is increasing and reaches stable when the sampling rate is 10\% for three sets of PolSAR data. So, we select 10\% as the training sample rate in the experiments.

\emph{3)Effect of the patch size in the network}

The patch size is  another parameter which effects the classification performance of the proposed method. In this paper, different patch sizes are selected to test the classification accuracies on three sets of PolSAR data covering Oberpfaffenhofen, Xi'an and San Francisco areas. The effect of patch size on classification accuracy for Oberpfaffenhofen, Xi'an and San Francisco data sets are presented in Table \ref{t8}, \ref{t9} and \ref{t10} respectively. The patch size is selected as $9\times9$, $13\times13$ and $17\times17$ to test the variations of classification accuracy respectively. It can be seen that the overall accuracy increases 2.45\% and 0.21\% respectively for Oberpfaffenhofen data set when patch size is enlarged from $9\times9$ to $17\times17$. For Xi'an data set, the overall accuracy increases 0.56\% and 0.26\%  respectively. For San Francisco data set,  the overall accuracy increases 0.06\% and 0.02\% respectively from $9\times9$ to $17\times17$.  It can be seen that the patch size with $13\times13$ can obtain more improvement in OA, and the patch with $17\times17$ has similar accuracy with $13\times13$. Therefore, we select $13\times13$ as the patch size in the proposed method.

\begin{table}[ht]
\footnotesize
\begin{center}
\caption
{ \label{t8}
 Classification accuracy of different patch size on Oberpfaffenhofen Data Set(\%).}
\begin{tabular}{p{1.8cm}p{1.4cm}p{1.3cm}p{1.3cm}}
\hline
&$9\times9$&$13\times13$&$17\times17$\\
\hline
farmland&95.82&97.07&\textbf{97.23}\\
suburban&94.57&97.58&\textbf{97.70}\\
woodland&93.38&97.84&\textbf{98.33}\\
road&93.74&\textbf{97.68}&97.28\\
open area&90.13&93.24&\textbf{93.75}\\
OA&94.43&96.88&\textbf{97.09}\\
AA&93.53&96.68&\textbf{96.86}\\
Kappa&91.88&95.46&\textbf{95.77}\\
\hline
\end{tabular}
\end{center}
\end{table}

\begin{table}[ht]
\footnotesize
\begin{center}
\caption
{ \label{t9}
 Classification accuracy of different patch size on Xi'an Data Set (\%).}
\begin{tabular}{p{1.8cm}p{1.4cm}p{1.3cm}p{1.3cm}}
\hline
&$9\times9$&$13\times13$&$17\times17$\\
\hline
water&\textbf{97.13}&95.18&94.60\\
grass&96.13&97.42&\textbf{98.39}\\
buildings&97.99&\textbf{98.59}&98.22\\
OA&96.94&97.50&\textbf{97.76}\\
AA&\textbf{97.08}&97.06&97.07\\
Kappa&94.97&95.87&\textbf{96.30}\\
\hline
\end{tabular}
\end{center}
\end{table}

\begin{table}[ht]
\footnotesize
\begin{center}
\caption
{ \label{t10}
 Classification accuracy of different patch size on San Francisco Data Set (\%).}
\begin{tabular}{p{1.8cm}p{1.4cm}p{1.3cm}p{1.3cm}}
\hline
&$9\times9$&$13\times13$&$17\times17$\\
\hline
ocean&100&100&100\\
vegetation&99.54&99.83&\textbf{99.86}\\
low-density&99.90&99.87&\textbf{99.91}\\
high-density&99.76&99.94&\textbf{99.96}\\
developed&99.62&99.83&\textbf{99.88}\\
OA&99.87&99.93&\textbf{99.95}\\
AA&99.76&99.89&\textbf{99.92}\\
Kappa&99.81&99.90&\textbf{99.93}\\
\hline
\end{tabular}
\end{center}
\end{table}

\emph{4)Analysis of running time}

We take the Xi'an data set as an example. The running time of the compared and proposed methods are given in Table \ref{time}, which lists the training and test time respectively. It can be seen the most time is costed in the training stage. The MPCNN method costs unbearable longer time than other methods. The proposed method achieves superior classification performance and takes less time than four compared methods. Compared with the 9D-CNN method, the proposed method takes more time on RCM module since it has many complex matrix operations. It proves the effectiveness of the proposed method in terms of time and performance.

\begin{table*}[ht]
\footnotesize
\begin{center}
\caption
{ \label{time}
 Computing time of different methods on Xi'an Data Set ($s$)}
\begin{tabular}{p{1.6cm}p{1.2cm}p{1.2cm}p{1.2cm}p{1.2cm}p{1.2cm}p{1.6cm}}
\hline
&CVCNN&DFGCN&CEGCN&MPCNN&9D-CNN&RCM\_CNN\\
\hline
training time&3498.4&960.52&830.19&21600.35&72.88&175.52\\
test time&17.45&2.59&4.03&327.53&5.83&7.68\\
\hline
\end{tabular}
\end{center}
\end{table*}

\section{Conclusion}

In this paper, we propose a novel Riemannian complex matrix convolution network(RCM\_CNN), which can effectively learn the geometric structure of complex matrix in the Riemannian space. It breaks through the Euclidian constraint for conventional convolution network, and develops a novel complex matrix network in Riemannian space for PolSAR images for the first time. In addition, the proposed RCM\_CNN method consists two modules, the RCMnet module is utilized to learn geometric structure and channel correlation of complex matrix in Riemannian space, and then the CNN module can learn the contextual high-level features based on the Riemannian features.  Experiments are conducted on three sets of real PolSAR data sets, and results demonstrate the proposed method can achieve superior classification performance than four state-of-the-art methods.

In addition, the proposed method explores a Riemannian complex matrix network structure followed by a CNN model to learn contextual information. In the further work, we try
to explore how to reduce computational time during using the back-propagation.

\section*{Acknowledgments}

This work was supported in part by the National Natural Science Foundation of China under Grant 62006186,62272383, the Science and Technology Program of Beilin District in Xi'an under Grant GX2105,
in part by the Open fund of National Key Laboratory of Geographic Information Engineering under Grant SKLGIE2019-M-3-2.

\bibliographystyle{IEEEtran}
\bibliography{mybibfile}

\end{document}